\documentclass[lettersize,journal]{IEEEtran}
\usepackage{amsmath,amssymb,amsfonts,pifont}
\usepackage{hyperref,graphicx}
\usepackage{tabu,float,textcomp}
\usepackage{multirow,dcolumn,hhline}
\usepackage{color,booktabs,relsize}
\usepackage[subrefformat=parens,labelformat=parens,caption=false]{subfig}
\usepackage{adjustbox}
\usepackage{gensymb}

\def\BibTeX{{\rm B\kern-.05em{\sc i\kern-.025em b}\kern-.08em
    T\kern-.1667em\lower.7ex\hbox{E}\kern-.125emX}}
\begin{document}
\title{Fusion of Quadratic Time--Frequency Analysis and Convolutional Neural Networks to Diagnose Bearing Faults Under Time-Varying Speeds}

\author{
Mohammad Al-Sa'd, \IEEEmembership{Senior Member, IEEE},
Tuomas Jalonen,
Serkan Kiranyaz, \IEEEmembership{Senior Member, IEEE},
and Moncef Gabbouj, \IEEEmembership{Fellow, IEEE}

\thanks{This work was supported by Business Finland Amalia and NSF IUCRC CBL project, and the Jenny and Antti Wihuri Foundation. \textit{(Corresponding author: Mohammad Al-Sa'd.)}}
\thanks{Mohammad Al-Sa'd, Tuomas Jalonen, and Moncef Gabbouj are with the Faculty of Information Technology and Communication Sciences, Tampere University, 33720 Tampere, Finland (e-mail:
\href{mailto:mohammad.al-sad@tuni.fi}{mohammad.al-sad@tuni.fi};
\href{mailto:tuomas.jalonen@tuni.fi}{tuomas.jalonen@tuni.fi}; \href{mailto:moncef.gabbouj@tuni.fi}{moncef.gabbouj@tuni.fi}).}
\thanks{Mohammad Al-Sa'd is also with the Faculty of Medicine and the Department of Clinical Neurophysiology at the University of Helsinki and Helsinki University Hospital, 00014, Helsinki, Finland (e-mail: \href{mailto:mohammad.al-sad@helsinki.fi}{mohammad.al-sad@helsinki.fi}; \href{mailto:ext-mohammad.al-sad@hus.fi}{ext-mohammad.al-sad@hus.fi}).}
\thanks{Serkan Kiranyaz is with the Department of Electrical Engineering, Qatar University, 2713 Doha, Qatar (e-mail: \href{mailto:mkiranyaz@qu.edu.qa}{mkiranyaz@qu.edu.qa}).}
}
\markboth{}
{\textit{Al-Sa'd}\MakeLowercase{\textit{ et al.}}: Fusion of Quadratic Time--Frequency Analysis and CNNs to Diagnose Bearing Faults Under Time-Varying Speeds}
\maketitle
\begin{abstract} 
Diagnosis of bearing faults is paramount to reducing maintenance costs and operational breakdowns. Bearing faults are primary contributors to machine vibrations, and analyzing their signal morphology offers insights into their health status. Unfortunately, existing approaches are optimized for controlled environments, neglecting realistic conditions such as time-varying rotational speeds and the vibration's non-stationary nature.
This paper presents a fusion of time-frequency analysis and deep learning techniques to diagnose bearing faults under time-varying speeds and varying noise levels. First, we formulate the bearing fault-induced vibrations and discuss the link between their non-stationarity and the bearing's inherent and operational parameters. We also elucidate quadratic time-frequency distributions and validate their effectiveness in resolving distinctive dynamic patterns associated with different bearing faults. Based on this, we design a time-frequency convolutional neural network (TF-CNN) to diagnose various faults in rolling-element bearings.
Our experimental findings undeniably demonstrate the superior performance of TF-CNN in comparison to recently developed techniques. They also assert its versatility in capturing fault-relevant non-stationary features that couple with speed changes and show its exceptional resilience to noise, consistently surpassing competing methods across various signal-to-noise ratios and performance metrics. Altogether, the TF-CNN achieves substantial accuracy improvements up to 15\%, in severe noise conditions.
\end{abstract}
\section{Introduction}
\IEEEPARstart{R}{otating} machinery, such as motors, gearboxes, wind turbines, generators, and engines, is at the heart of modern industrial applications \cite{alshorman2020review}. Condition monitoring and fault diagnosis of rotating machinery allow proactive maintenance strategies and reduce economic and operational impacts caused by unexpected failures \cite{rai2016review}. Bearing faults constitute a prominent category of irregularities in rotating machines; thus, continuous monitoring is imperative \cite{rai2016review,DeekshitKompella2018}. Fortunately, abnormal machine vibrations are largely a result of faulty bearings, and analyzing their characteristics facilitates diagnosing the bearing's health condition \cite{jalonen2023bearing_fault,9558835,youcef2020rolling,Malla2019}.

Deep learning techniques, such as Convolutional Neural Networks (CNNs), are adequate to diagnose the bearing's condition by detecting irregularities in the vibration data \cite{7501527,Eren2019,8584489,jalonen2023realtime}. Nevertheless, prevailing solutions are usually tested under ideal conditions and often confined to fixed \cite{Jiang2023,Prudhom2017,Zhang2019} or slowly increasing/decreasing rotational speeds \cite{Attoui2017,9247131} that do not represent dynamic scenarios \cite{jalonen2023bearing_fault}, i.e., time-varying changes. The vibratory profile of bearings under time-varying rotational speeds is strictly non-stationary \cite{Feng2013}, i.e., vibration is modulated by speed and characterized by a time-varying spectral content \cite{TFBook,Zhang2022}. Therefore, classical time- or frequency-domain representations are insufficient, and one must adopt time-frequency (TF) tools to explore the signal's degrees of freedom \cite{4524002}. 
Therefore, in this paper, we propose to leverage synergies between TF analysis and deep learning to diagnose bearing faults under highly time-varying rotational speeds. The key contributions of this paper are:
\begin{itemize}
    \item Quadratic time-frequency distributions are effective for revealing distinctive dynamic patterns that are associated with different bearing faults.
    \item A bearing fault diagnosis system fusing quadratic time-frequency distributions and CNNs with robustness to diverse noise levels and time-varying speed conditions.
    \item Precise identification of various bearing faults with performance significantly surpassing recently developed techniques about 15\% gain in average accuracy.
\end{itemize}

Time-frequency representations (TFR) fall under two complementary categories; linear and quadratic \cite{TFBook}. Linear TFRs, such as short-time Fourier transform and continuous wavelet transform, extend the signal's spectrum to become time-dependent \cite{TFBook}. Specifically, they use a set of basis functions for decomposition and yield time-varying complex-valued frequency representations \cite{al2022time}. The resolution of linear TFRs is limited by the Heisenberg-Gabor uncertainty principle \cite{al2022time,6633061}. However, quadratic TFRs, derived from the Wigner-Ville distribution (WVD), overcome this limitation by transforming the signal's time-varying auto-correlation function \cite{TFBook}. Nevertheless, this advantage is at the expense of reducing accuracy by introducing artifactual patterns, named cross-terms, that require filtering \cite{9456035}. Fortunately, high-performing distributions are designed to achieve the best trade-off between accuracy and resolution \cite{9456035}.

The utility of linear TFRs was found useful for analyzing various faults in rolling-element bearings \cite{10138769,Cai_2017,Wang2019} and for classifying their health states \cite{Zhang2020,9094580,kaya2022new}. Nonetheless, most experiments were conducted with limited speed profiles and were confined to constant \cite{kaya2022new,9167237}, fixed levels \cite{9094580}, or slowly varying speeds with limited range \cite{Wang2023}. Besides, some studies focused on analyzing the extracted features rather than reporting validated classification results \cite{10138769,Wang2019,8676242}.
Moreover, quadratic TFRs were utilized in only a few studies using elementary distributions \cite{Quinde2019,8675278,Dong2012}, although there are better high-performing alternatives such as the compact kernel distribution, smoothed-pseudo Wigner-Ville distribution, and the extended modified B-distribution \cite{9456035}. In fact, a recent review reveals that the literature is dominantly focused on developing post-processing techniques for quadratic TFRs rather than utilizing more advanced representations \cite{10285399}.
Consequently, there is a need for fusing high-performing quadratic TFRs with deep learning techniques to build a precise tool for diagnosing non-stationary bearing faults.
\section{Methodology} \label{sec:methodology}
\begin{figure}[!t]
\centerline{\includegraphics[width=0.3\textwidth]{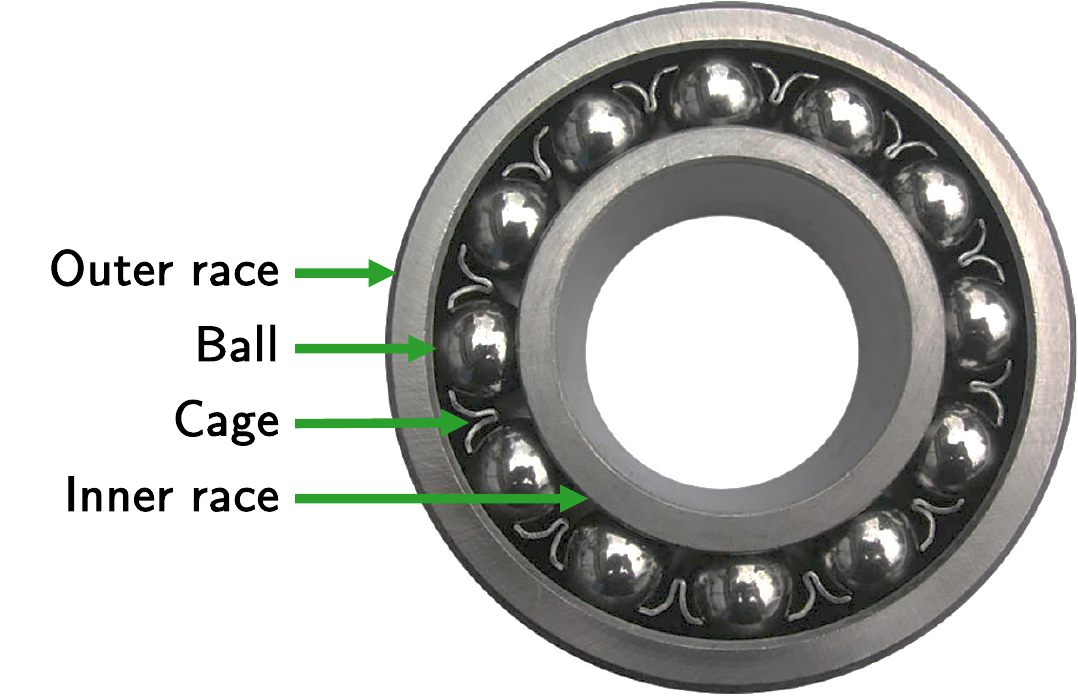}}
    \caption{A rolling element bearing example. The bearing's outer race, inner race, balls, and cage are illustrated. Redrawn from \cite{bearing_photo}.}
    \label{fig:bearing}
\end{figure}

The role of a ball bearing is to reduce friction and to provide support for both radial and axial loads \cite{Kumar2018}. This is achieved by enclosing the steel rollers with two races to enable load transmissions \cite{Kumar2018}; see Fig. \ref{fig:bearing}. Fatigue in bearings introduces spalling, a common type of damage that primarily occurs on the races and the rolling elements, and changes the machine's vibratory profile \cite{Sawalhi2011}. Under time-varying rotational speeds, this fault-induced alteration becomes dynamic and co-varies with the motor's speed. In other words, speed modulates the vibrations and makes them non-stationary, i.e., characterized by time-varying spectra. Therefore, we combined TF analysis and deep learning techniques to develop a precise tool for diagnosing non-stationary faults in bearings.
\subsection{Signal model}
The bearing vibrations, captured by accelerometers mounted on the horizontal and vertical axes of the bearing's housing, can be generally modeled by:
\begin{equation} \label{eq:signal_model}
    \boldsymbol{x}(t) = \boldsymbol{s}(t) + \alpha\,\boldsymbol{\eta}(t)\,,
\end{equation}
where $\boldsymbol{x}(t) = [x_{\text{x}}(t), x_{\text{y}}(t)]^T$ holds noisy vibrations recorded by the accelerometers mounted on the x and y directions, $\boldsymbol{s}(t)=[s_{\text{x}}(t), s_{\text{y}}(t)]^T$ contains the bearing's noise-free vibrations in both directions, $\boldsymbol{\eta}(t)=[\eta_{\text{x}}(t), \eta_{\text{y}}(t)]^T$ holds independent and identically distributed white Gaussian noise signals with zero mean and unit variance, and $\alpha = [\alpha_{\text{x}}, \alpha_{\text{y}}]^T$ are factors that control the amount of noise-induced degradation based on a predefined signal-to-noise ratio (SNR) in decibels (dB) and are calculated as follows:
\begin{equation}
    \alpha = \sqrt{10^{-\text{SNR}/10}\,\dfrac{1}{T} \int_0^T \boldsymbol{s}^2(t)\,\text{d}t}\,,
\end{equation}
where $T$ is the time duration of the signals in $\boldsymbol{s}(t)$.
The bearing noise-free vibrations are comprised of healthy patterns (random or deterministic) and fault-induced amplitude-modulated oscillations \cite{Kim2020,Feng2017,saruhan2014vibration}, i.e.:
\begin{equation}
    \boldsymbol{s}(t) = \boldsymbol{h}(t) + \boldsymbol{f}(t)\,,
\end{equation}
\small{
\begin{equation} \label{eq:fault_model}
    \boldsymbol{f}(t) = \underbrace{\vphantom{\delta\left(t-\dfrac{k}{f_{\text{fault}}(t)}\right)}A\sin(2\pi f_c t)}_{\text{Amplitude modulation}}\sum_{k=0}^{K-1}\underbrace{\vphantom{\delta\left(t-\dfrac{k}{f_{\text{fault}}(t)}\right)}e^{-\beta f_c t}}_{\text{Fault impulses}}\underset{(t)}{*} \underbrace{\delta\left(t-\dfrac{k}{f_{\text{fault}}(t)}\right)}_{\text{Pulse frequency modulation}}\,,
\end{equation}
}
where $\boldsymbol{h}(t)$ holds the bearing normal vibration (healthy patterns), $\boldsymbol{f}(t)$ is the fault-specific waveform, $A$ is the fault amplitude, $f_c$ is the bearing resonance frequency, $\beta$ is the damping characteristic factor, $\underset{(t)}{*}$ denotes convolution in time, $\delta(t)$ is the Dirac function, $f_{\text{fault}}(t)$ denotes the time-varying fault characteristic frequency, and $K$ is the number of fault impulses, or knocks, within the signal duration $T$. The characteristic frequencies that appear in the vibration signal of a faulty bearing, assuming a stationary outer race and a rotating inner race, are mainly the ball pass frequency of the inner race, ball pass frequency of the outer race, and ball spinning frequency, which are expressed by \cite{Cheng2019}:
\begin{equation} \label{eq:fault_1}
    f_{\text{fault}}^{(\text{Inner})}(t) = \dfrac{n}{2} \left(1+\dfrac{D_r}{D_p} \cos(\theta)\right)f_r(t)\,,
\end{equation}
\begin{equation} \label{eq:fault_2}
    f_{\text{fault}}^{(\text{Outer})}(t) = \dfrac{n}{2} \left(1-\dfrac{D_r}{D_p} \cos(\theta)\right)f_r(t)\,,
\end{equation}
\begin{equation} \label{eq:fault_3}
    f_{\text{fault}}^{(\text{Ball})}(t) = \dfrac{D_p}{D_r} \left(1-\left(\dfrac{D_r}{D_p} \cos(\theta)\right)^2\right)f_r(t)\,,
\end{equation}
where $n$ is the number of rolling elements, $D_r$ is the rolling element diameter, $D_p$ is the pitch diameter, $\theta$ denotes the contact angle, i.e., the angle of load from the radial plane, and $f_r(t)$ is the motor's time-varying rotational frequency, i.e., number of rotations per second ($\text{rpm}/60$).
\begin{figure*}[!t]
    \centering
    \subfloat[Time-varying rotational speed. \label{fig:model_speed}]{\includegraphics[width=.33\textwidth]{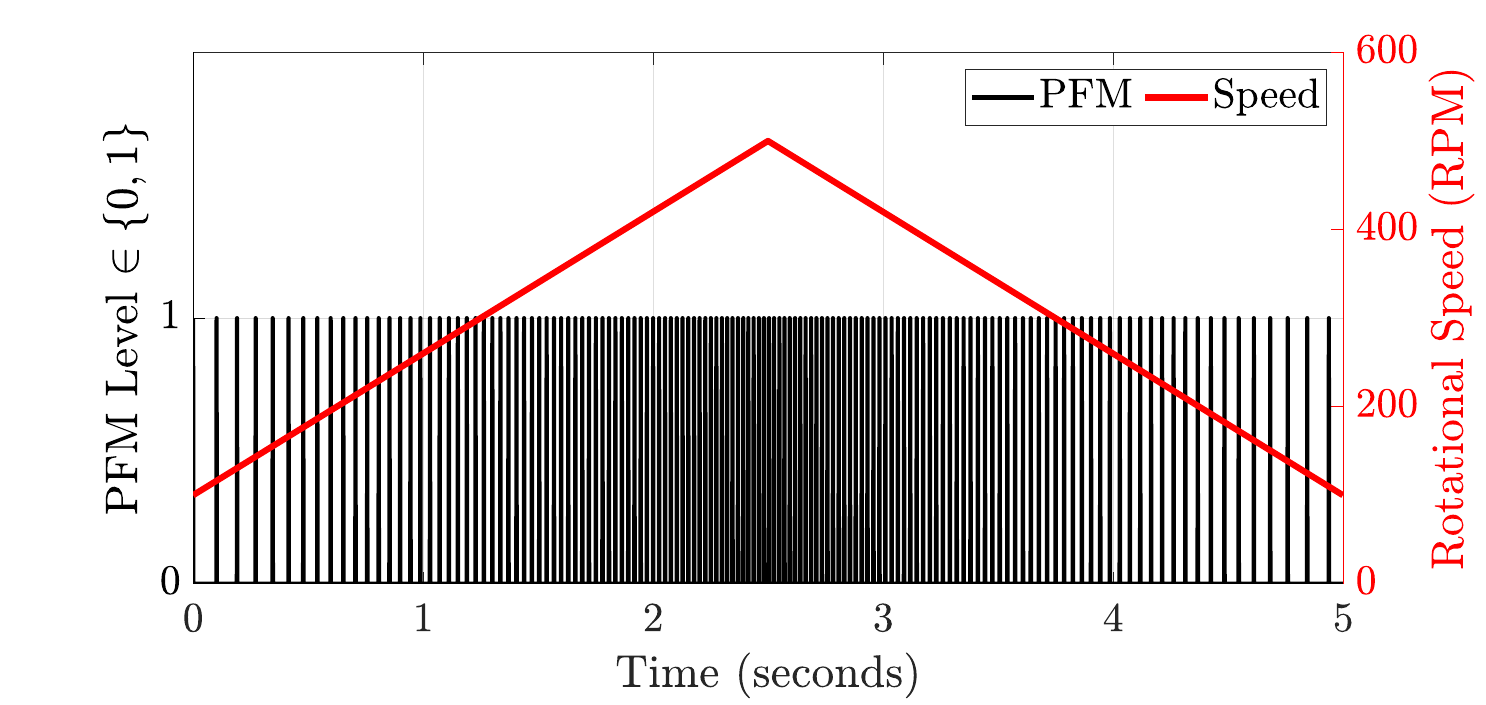}}
    \subfloat[The resultant vibration signal. \label{fig:model_vibration}]{\includegraphics[width=.33\textwidth]{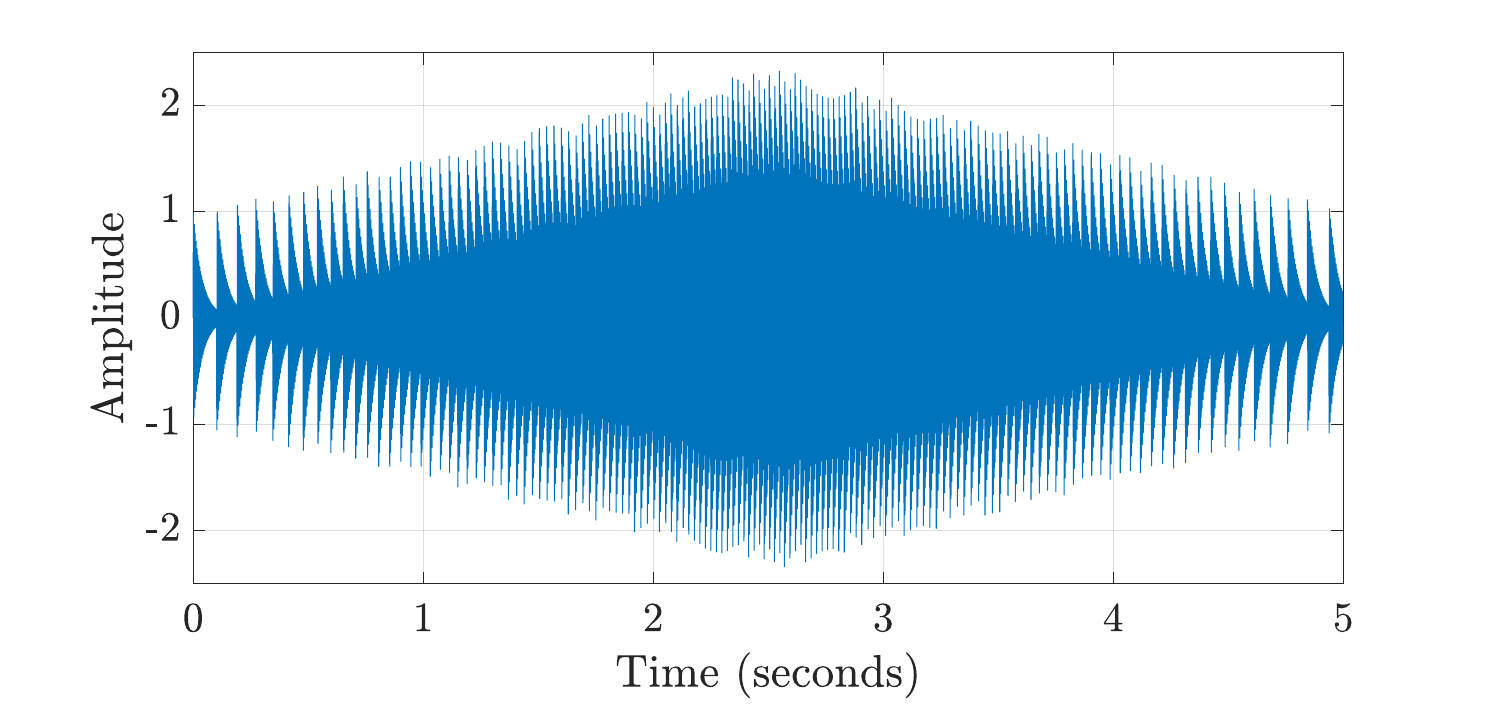}}
    \subfloat[The signal's frequency representation. \label{fig:model_spectra}]{\includegraphics[width=.33\textwidth]{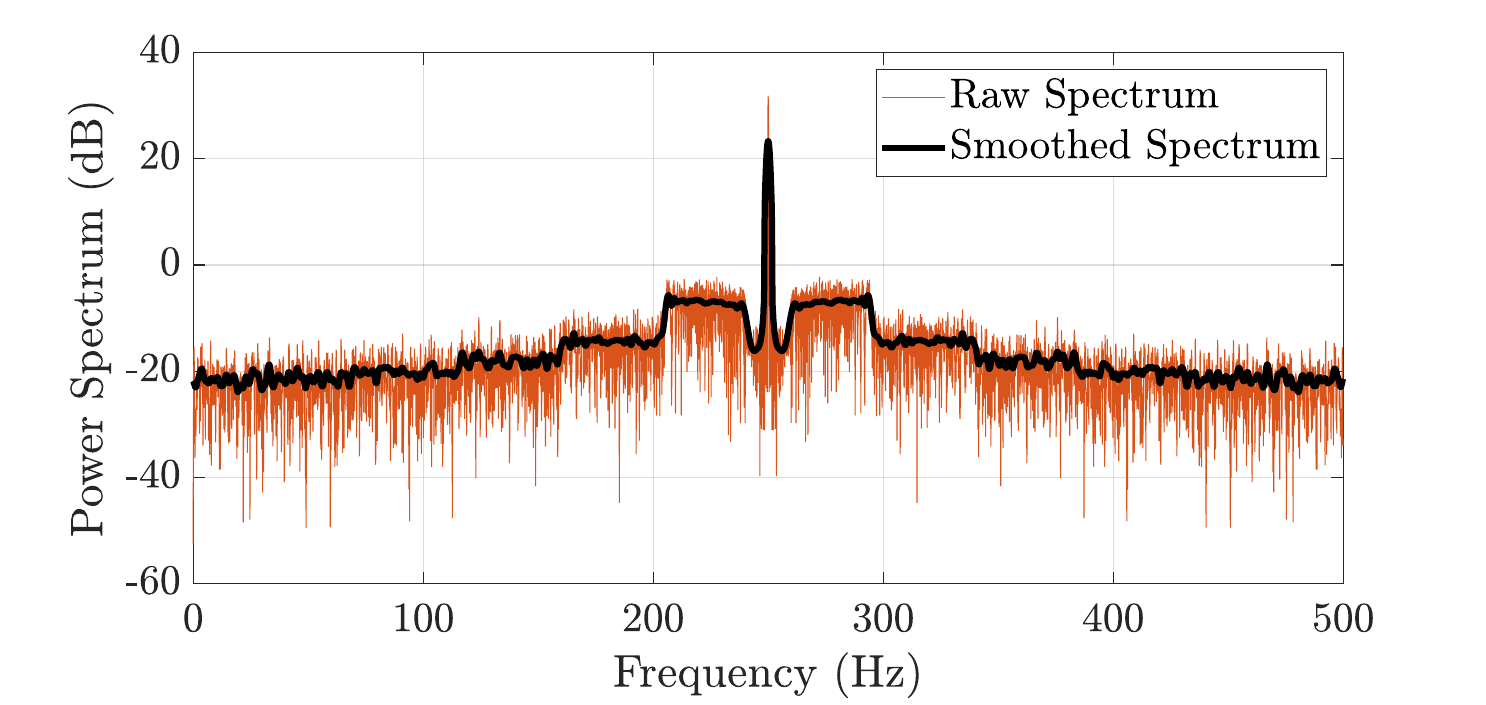}}
    \\
    \subfloat[The signal's WVD. \label{fig:model_wvd}]{\includegraphics[width=.33\textwidth]{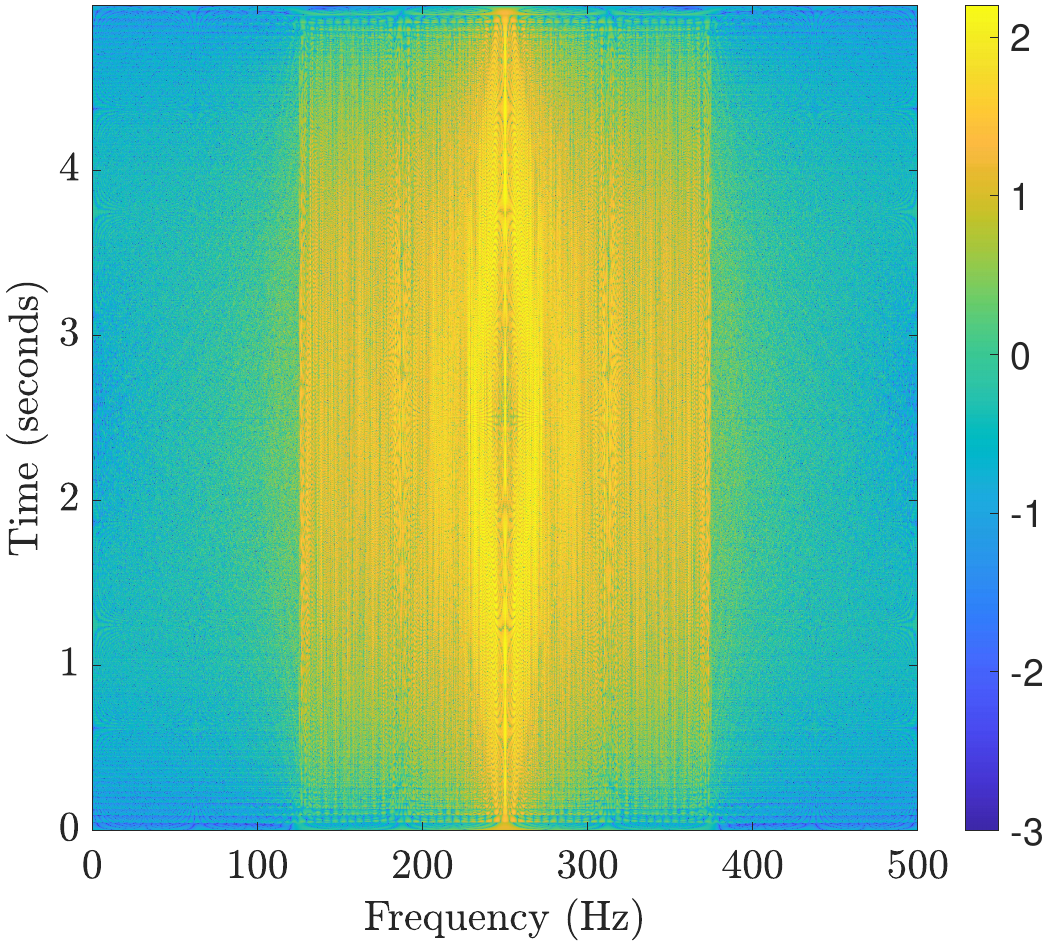}}
    \subfloat[The signal's CKD. \label{fig:model_ckd}]{\includegraphics[width=.33\textwidth]{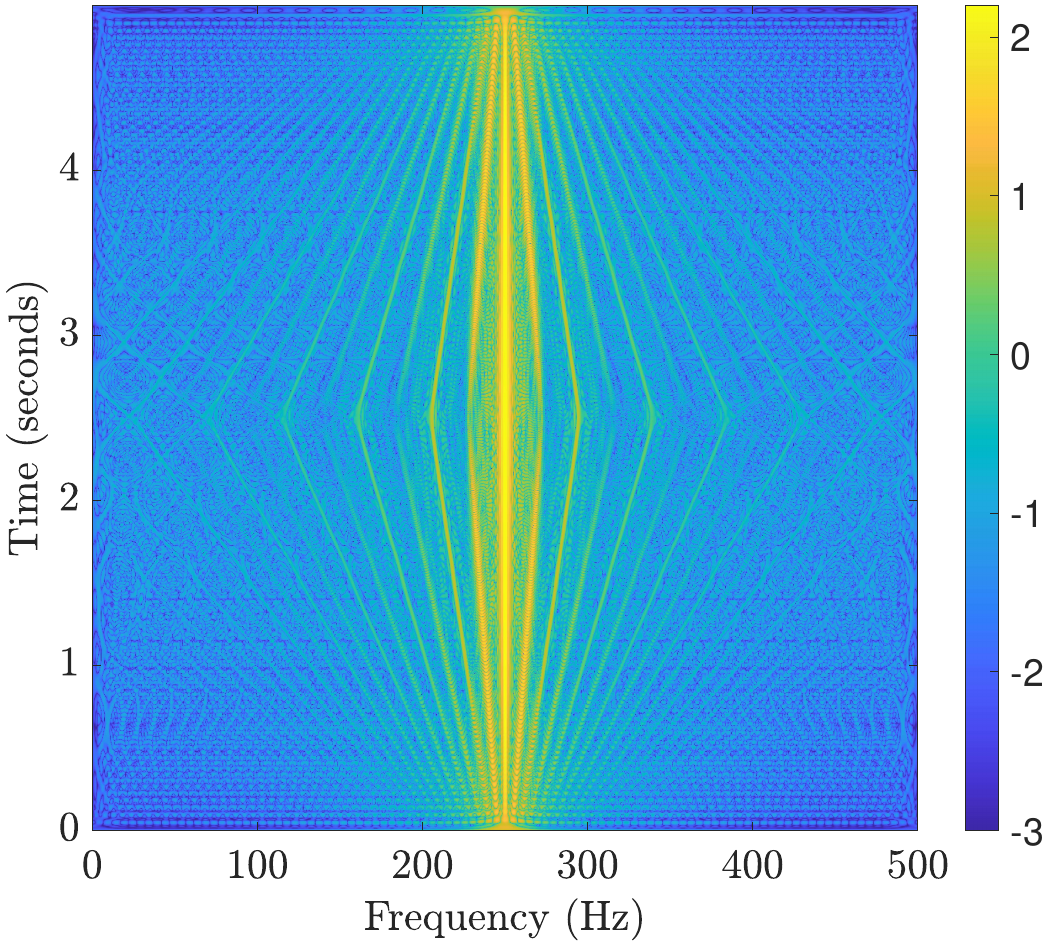}}  
    \subfloat[The signal's Spectrogram. \label{fig:model_spec}]{\includegraphics[width=.33\textwidth]{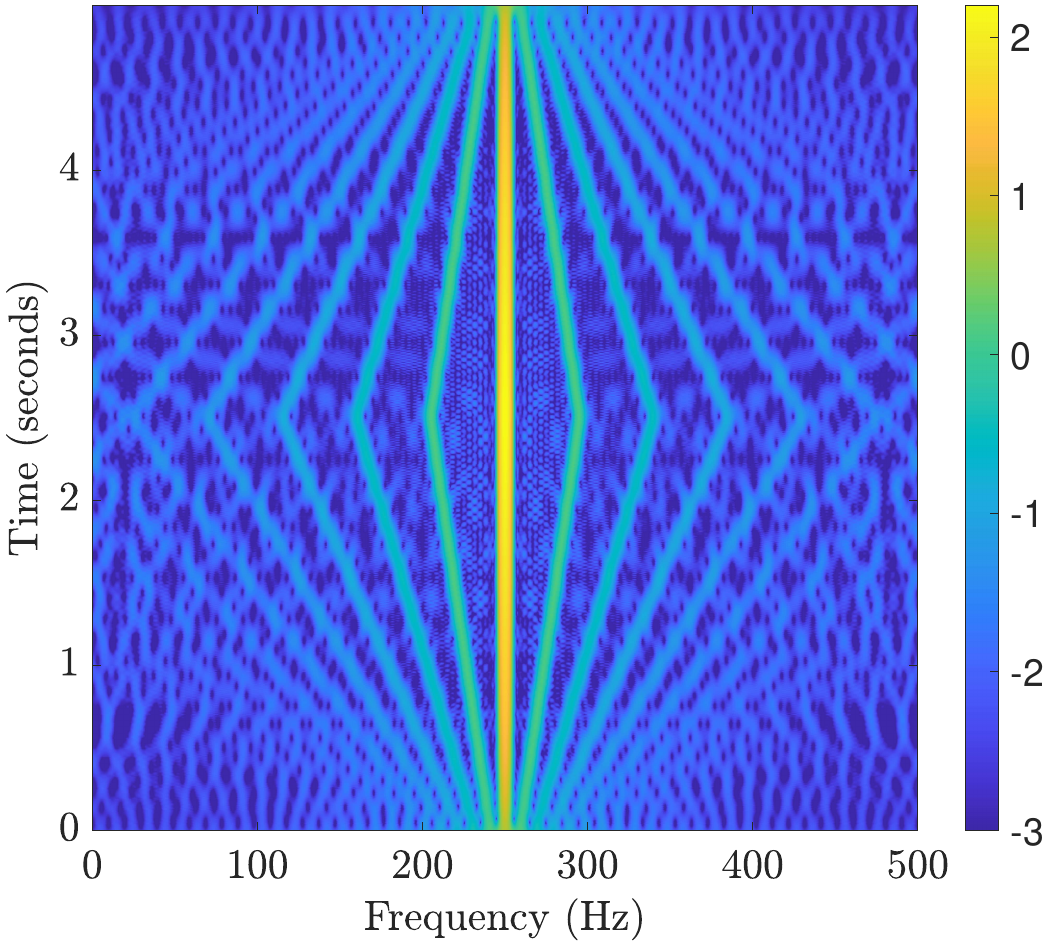}}
    \caption{Example synthetic bearing vibration signal with an inner race fault following Eq. (\ref{eq:fault_model}). (a) shows the linearly changing rotational speed in red and its pulse frequency modulation (PFM) encoding in black, (b) depicts the generated vibration signal, (c) illustrates its power spectrum, and (d)-(f) show the compact kernel distribution (CKD) high accuracy, compared to the Wigner-Ville distribution (WVD), and its high resolution, compared to the Spectrogram, when representing the vibration signal in the time-frequency domain. The signal model parameters are set to $f_s = 1$ kHz, $T = 5$ seconds, $A = 1$, $f_c = 250$ Hz, $\beta = 0.1$, $n = 9$, $D_r = 7.9$ mm, $D_p = 38.5$ mm, $\theta=0\degree$, and $f_r(t)$ is a triangular wave rising from 100 rpm to 500 rpm and back to 100 rpm. The CKD parameters are set to $c=1$, $D=0.1$, $E=0.1$, and the Spectrogram is computed at full resolution using a 0.3 seconds Hamming window. The TFRs are plotted on a logarithmic scale to ease visualizing the fault-induced waveforms around the bearing's resonance at $250$ Hz (-3 dB to 2.2 dB).}
   \label{fig:signal_model}
\end{figure*}

The bearing fault vibrations, expressed in Eq. (\ref{eq:fault_model}), are comprised of three parts: amplitude modulation (AM), fault-induced impulses, and pulse frequency modulation (PFM); see Figs. \ref{fig:model_speed}-\ref{fig:model_spectra} for an example. First, the PFM component encodes the fault time-varying frequency which, by examining Eqs. (\ref{eq:fault_1})-(\ref{eq:fault_3}), depends on the bearing's geometry, the fault location, the motor's rotational speed, and the contact angle. Specifically, the bearing's geometry and the location of the fault are responsible for defining the fault's base harmonics, the motor's rotational speed shifts the base spectrum in a time-dependent manner, and variations in the contact angle due to misalignment, thermal growth, excessive bolt tightening, and pitting/peeling in the bearing races induce further frequency disruptions \cite{Zhang2015}. Moreover, the fault impulses in Eq. (\ref{eq:fault_model}) express the rolling element rattlings that depend on the bearing geometry and emerge with the frequency defined by the PFM. Finally, the AM component carries the frequency-modulated impulses to the bearing's resonance frequency. Altogether, the vibration signal of a faulty bearing, under time-varying rotational speed conditions, is strictly non-stationary and mandates TF tools for analysis and processing.
\subsection{Quadratic time-frequency representation}
A time-frequency representation (TFR), obtained by a time-frequency distribution (TFD), describes the vibrations spectra through time \cite{TFBook}, i.e., it reveals the vibrations' joint temporal and spectral evolution \cite{9456035}.
Let $\boldsymbol{z}(t) = [z_{\text{x}}(t), z_{\text{y}}(t)]^T$ be the analytic associate of the vibration signals in $\boldsymbol{x}(t)$ obtained via the Hilbert transform, such that:
\begin{equation}
    \boldsymbol{z}(t) = \boldsymbol{x}(t) + j\mathcal{H}\left\{\boldsymbol{x}(t)\right\}\,,
\end{equation}
where $\mathcal{H}\left\{\,\cdot\,\right\}$ is the Hilbert transform and $j=\sqrt{-1}$.
The Wigner-Ville distribution (WVD) of $\boldsymbol{z}(t)$ is defined as \cite{9456035}:
\begin{equation} \label{eq:wvd}
    \boldsymbol{W}_{\!\!\boldsymbol{z}}(t,f) = \underset{\tau \rightarrow f}{\mathcal{F}} \bigg\{\boldsymbol{z}\left(t+\dfrac{\tau}{2}\right) \boldsymbol{z}^*\left(t-\dfrac{\tau}{2}\right)\bigg\}\,,
\end{equation}
where $\boldsymbol{W}_{\!\!\boldsymbol{z}}(t,f)$ holds the WVDs of the signals in $\boldsymbol{z}(t)$, $\underset{\tau \rightarrow f}{\mathcal{F}}\left\{\,\cdot\,\right\}$ denotes the Fourier transform (FT) from lag $\tau$ to frequency $f$, and $\boldsymbol{z}^*(t)$ is the complex conjugate of $\boldsymbol{z}(t)$ \cite{Boashash2018}.
The WVD is optimal for mono-component linear frequency modulated signals, but when analyzing other signal types, it is known to suffer from interference patterns, named cross-terms, because of its bilinear nature \cite{9456035}. The cross-terms severity can be reduced by a TF smoothing kernel $\gamma(t,f)$ such that:
\begin{equation} \label{eq:htfd}
    \boldsymbol{\rho}_{\boldsymbol{z}}(t,f) = \gamma(t,f)\underset{(t,f)}{**}\boldsymbol{W}_{\!\!\boldsymbol{z}}(t,f)\,,
\end{equation}
where $\boldsymbol{\rho}_{\boldsymbol{z}}(t,f)$ holds smoothed TFDs and $\underset{(t,f)}{**}$ denotes the convolution in time and frequency.
This expression can be written in a more convenient space, called the Doppler-lag domain, where convolutions become multiplications \cite{TFBook}, i.e.:
\begin{equation}\label{eq:tfd}
    \boldsymbol{\rho}_{\boldsymbol{z}}(t,f) = 
    \underset{t \leftarrow \nu}{\mathcal{F}}^{-1} \bigg\{ \underset{\tau \rightarrow f}{\mathcal{F}} \bigg\{g(\nu,\tau)  \boldsymbol{A}_{\boldsymbol{z}}(\nu,\tau)\bigg\}\bigg\}\,,
\end{equation}
\begin{equation}
    g(\nu,\tau) = \underset{\tau \leftarrow f}{\mathcal{F}}^{-1} \bigg\{ \underset{t \rightarrow \nu}{\mathcal{F}} \bigg\{\gamma(t,f)\bigg\}\bigg\}\,,
\end{equation}
\begin{equation} \label{eq:af}
    \boldsymbol{A}_{\boldsymbol{z}}(\nu,\tau) = \underset{t \rightarrow \nu}{\mathcal{F}} \bigg\{\boldsymbol{z}\left(t+\dfrac{\tau}{2}\right) \boldsymbol{z}^*\left(t-\dfrac{\tau}{2}\right)\bigg\}\,,
\end{equation}
where $\underset{t \leftarrow \nu}{\mathcal{F}}^{-1}\left\{\,\cdot\,\right\}$ denotes the inverse FT from Doppler $\nu$ to time $t$, $g(\nu,\tau)$ is the Doppler-lag kernel, and $\boldsymbol{A}_{\boldsymbol{z}}(\nu,\tau)$ is the ambiguity function (AF).
The Doppler-lag kernel's role is to mitigate the WVD cross-terms while concurrently preserving the resolution of its auto-terms, which encapsulate the genuine content \cite{al2022time}. Employing a compact kernel distribution (CKD) stands out as a highly effective method for accomplishing this critical task \cite{9456035}. Besides, it is more suitable for resolving the vibration waveforms of faulty bearings in the TF domain when compared to the WVD and Spectrogram; see Figs. \ref{fig:model_wvd}-\ref{fig:model_spec}.
The CKD utilizes a separable compact support kernel that is comprised of two parts operating independently in the Doppler-lag domain such that:
\begin{equation} \label{eq:sep_kernel}
    g(\nu,\tau) = G(\nu) \, \mathcal{G}(\tau)\,,
\end{equation}
\begin{equation}
    G(\nu) = \exp\left(c+\frac{c\,D^2}{\nu^2-D^2}\right) : |\nu|<D\,,
\end{equation}
\begin{equation}
    \mathcal{G}(\tau) = \exp\left(c+\frac{c\,E^2}{\tau^2-E^2}\right) : |\tau|<E\,,
\end{equation}
where $G(\nu)$ and $\mathcal{G}(\tau)$ are the independent Doppler and lag kernels, respectively, $D\in[0,1]$ and $E\in[0,1]$ are the kernel normalized cut-offs along the Doppler and lag axes, respectively, and $c>0$ defines the shape of the kernel.
\subsection{Bearing fault diagnosis}
\subsubsection{Dataset description}
This study utilizes an open-access dataset collected at the Korean Advanced Institute of Science and Technology (KAIST) \cite{KAIST_data}. The dataset focuses on diagnosing bearing faults under time-varying motor speeds. The conducted experiments include vibration signals captured at 25.6 kHz by accelerometers installed on the x- and y-axes of a bearing housing \cite{KAIST_data}. Faults were induced in the bearing, and vibration data was recorded for four distinct health states: \emph{Normal}, \emph{Outer}, \emph{Inner}, and \emph{Ball}, delineating a typical bearing function and three common faults. In total, 35 minutes of multi-sensor data was collected for each of the four classes.
\subsubsection{Data processing and representation}
\begin{figure}[!t]
    \centerline{\includegraphics[width=0.45\textwidth]{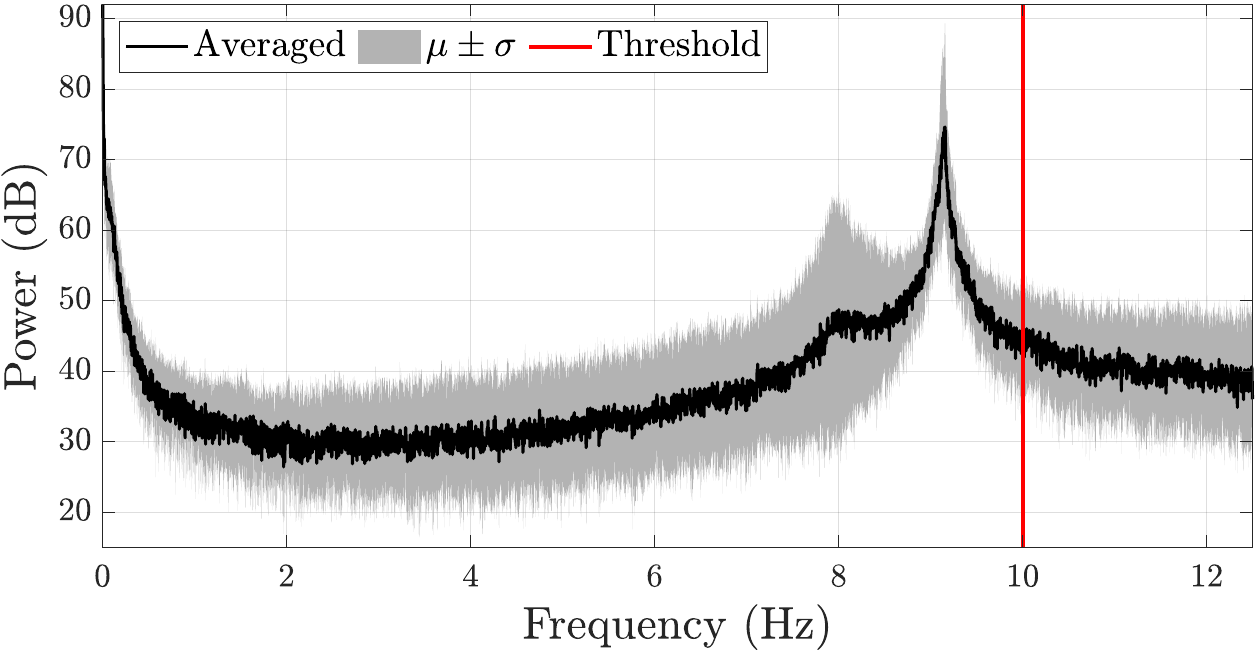}}
    \caption{Frequency analysis of the rotational speeds in the KAIST dataset. The motor speed exhibits variations at two distinct frequencies: 8 Hz and 9.15 Hz, with a maximum rate of change of 10 Hz as it includes 99.6\% of the total power spectrum, covering nearly all rapid speed changes. The power spectrum is estimated by the nonuniform FT because of the speed's inconsistent acquisition rate, averaged across the classes and dataset files, and plotted along with its 68.3\% confidence interval. Frequency scaling uses a 12.5 Hz sampling frequency which is the reciprocal of the shortest acquisition time.}
    \label{fig:segment_length}
\end{figure}

\begin{figure*}[!t]
    \centering
    \subfloat[Normal vibrations (\emph{Normal}). \label{fig:ckd1}]{\includegraphics[width=.25\textwidth]{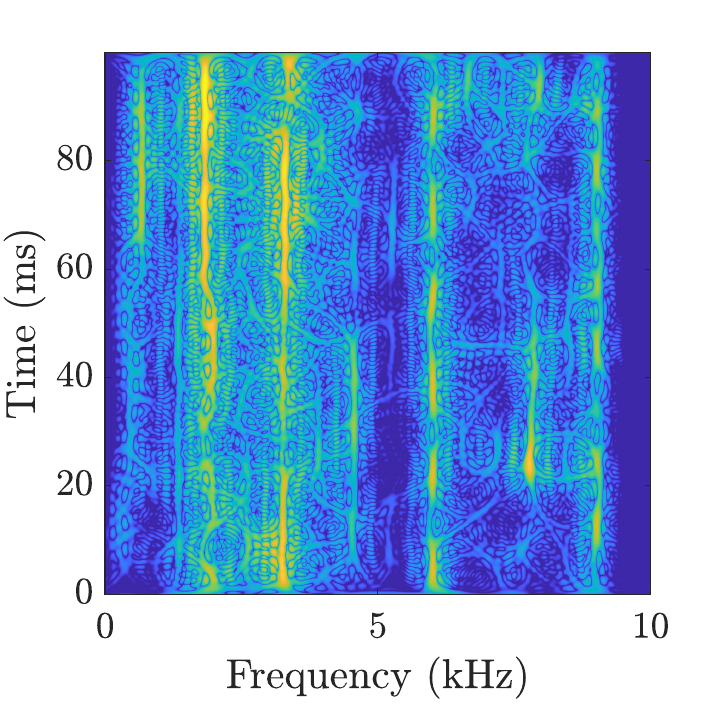}}
    \subfloat[Inner race fault (\emph{Inner}). \label{fig:ckd2}]{\includegraphics[width=.25\textwidth]{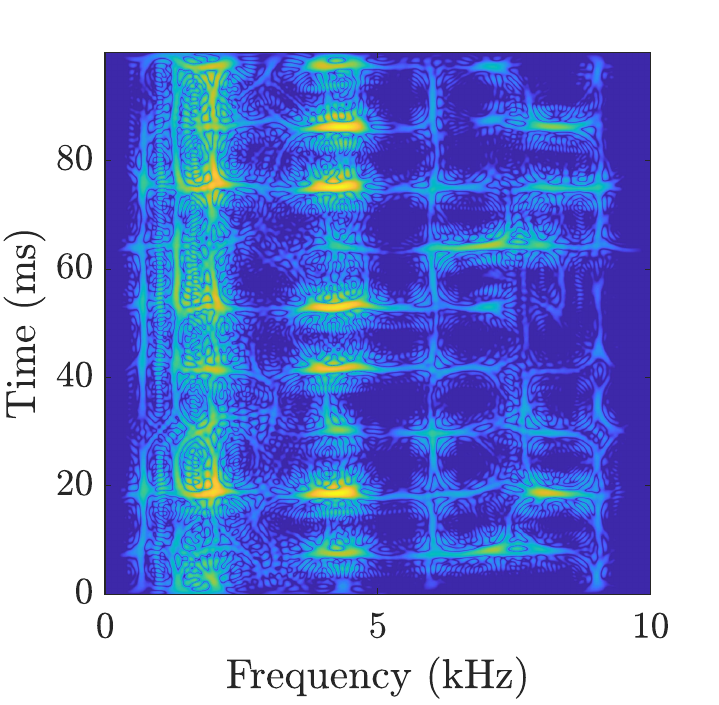}}
    \subfloat[Outer race fault (\emph{Outer}). \label{fig:ckd3}]{\includegraphics[width=.25\textwidth]{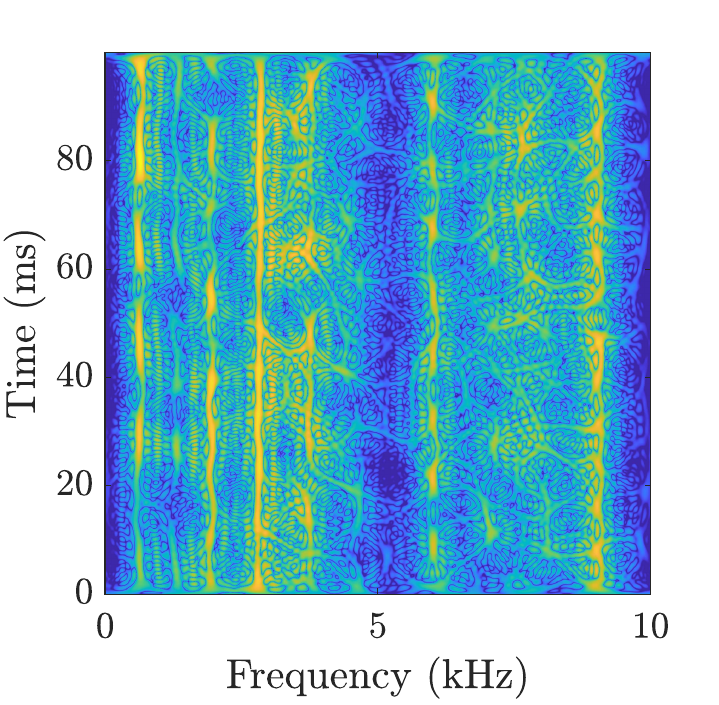}}
    \subfloat[Bearing ball fault (\emph{Ball}). \label{fig:ckd4}]{\includegraphics[width=.25\textwidth]{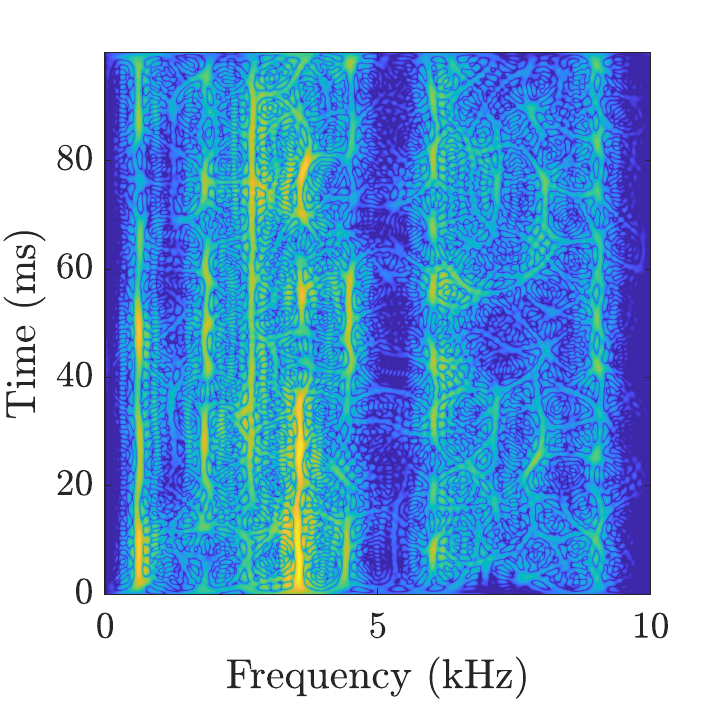}}
    \caption{The CKD of the first vibration segment from the sensor mounted on the y-direction with no noise. The representations' magnitudes are plotted on a logarithmic scale to ease visualization (-3 dB to 0 dB).}
    \label{fig:tf_example}
\end{figure*}

The accelerometer waveforms were first downsampled to 20 kHz using a finite impulse response anti-aliasing lowpass filter with delay compensation, which resulted in 42M samples for each sensor and bearing state.
After that, we divided the signals' temporal course into smaller segments with no overlap to prevent any unintended data leakage when training and/or testing the fault diagnosis system. However, selecting an appropriate segment length requires careful consideration. On one hand, it must be sufficiently small to provide an ample number of samples for training while minimizing additional complexities stemming from the motor's variable speed. On the other hand, it should be broad enough to encompass intricate patterns and capture class-specific features. We decided 0.1 seconds was suitable because the motor speed changes at a maximum rate of 10 Hz, as illustrated by the frequency analysis in Fig. \ref{fig:segment_length}. This segment length yields 21K segments for each sensor and bearing state.
Moreover, we generated the TFR of each segment as follows: (1) form the signal's AF by Eq. (\ref{eq:af}); (2) multiply the AF with the compact support kernel ($c=1$, $D=0.1$, and $E=0.1$) to suppress cross-terms; (3) revert the filtered AF to the TF domain using Eq. (\ref{eq:tfd}); (4) downsample the resultant TFR from $2000\times2000$ to $128 \times 128$ to reduce computational complexity (see Fig. \ref{fig:tf_example} which demonstrates the CKD of an example segment for all the bearing health states); and (5) standardize the TFR to have zero mean and unit variance.
\subsubsection{Time-frequency Convolutional Neural Network (TF-CNN)}
We developed a lightweight CNN architecture in a systematic manner, named TF-CNN, that accepts the standardized TFRs coming from both accelerometers (input size: $128\times128\times2$). The input TFRs undergo processing through five convolutional blocks, each comprising a convolutional layer for feature extraction employing a ReLU activation function to address non-linear effects, and a max-pooling layer of size $2\times2$ for summarizing the features. The convolutional layer in the first block has a kernel size of $5\times5$, while subsequent layers utilize kernels sized at $3\times3$. The features extracted from the final convolutional block are flattened and directed through a dense layer housing $128$ nodes activated by ReLU. Subsequently, they traverse a dropout layer with $0.4$ rate and a classification layer comprising $4$ nodes with a Softmax activation function to transform the network's outputs to probabilities.
\subsubsection{Experiment settings}
We trained and tested the TF-CNN model in a 5-fold cross-validation fashion to accurately assess its capacity for generalization. Specifically, the 84K samples (four classes each containing 21K segments) were partitioned into five equally sized and shuffled stratified folds. Within this process, one fold served as the test set, while the remaining four were utilized for training. This procedure was iterated five times, and the training process extended across 150 epochs using an Adam optimizer to minimize categorical cross-entropy. We set the batch size to 100 and the learning rate to $10^{-5}$ for epochs 0 to 100, followed by a reduced rate of $10^{-6}$ for the rest of the epochs. Finally, we tested the model using the weights learned at the last epoch (no early stopping).
In this study, we experimented with clean vibration signals and those degraded by noise at 5 dB, 0 dB, and -5 dB SNRs. Despite the varying noise levels, we utilized the same TF-CNN architecture and training configurations. Additionally, we compared the TF-CNN to two techniques recently developed for the KAIST dataset; PIResNet \cite{Ni2023} and an efficient time-domain CNN (T-CNN) \cite{jalonen2023bearing_fault}. These comparisons are largely for testing the value of representing the vibration signals in the TF domain via quadratic TFDs.
\subsubsection{Performance evaluation and analysis}
We evaluated the diagnosis of multiple bearing faults in a one-versus-all fashion, converting the original multi-class problem into a series of binary classification tasks. We quantified the models' efficacy in each task by accuracy, precision, recall, and F1-score, and then averaged these measures across all tasks.
Moreover, the TF-CNN was assessed using t-distributed stochastic neighbor embedding (t-SNE) and Gradient-weighted Class Activation Mapping (Grad-CAM) to allow an understanding of its predictions when presented with new vibration signals. Specifically, t-SNE reduces data dimensionality by grouping similar samples and separating dissimilar ones \cite{van2008visualizing}, and Grad-CAM generates a distribution with high values for pixels that contributed more to the network's decision \cite{selvaraju2017grad}. In general, with a sufficiently extensive training set, if these techniques show genuine learning, one may infer the model's adequacy for unseen samples.

\section{Results and discussion} \label{sec:results}
\subsection{Evaluation of bearing fault diagnosis}
\begin{figure*}[!t]
   \centering
   \subfloat[Clean data (no added noise). \label{fig:cm_1}]{\includegraphics[width=.24\textwidth]{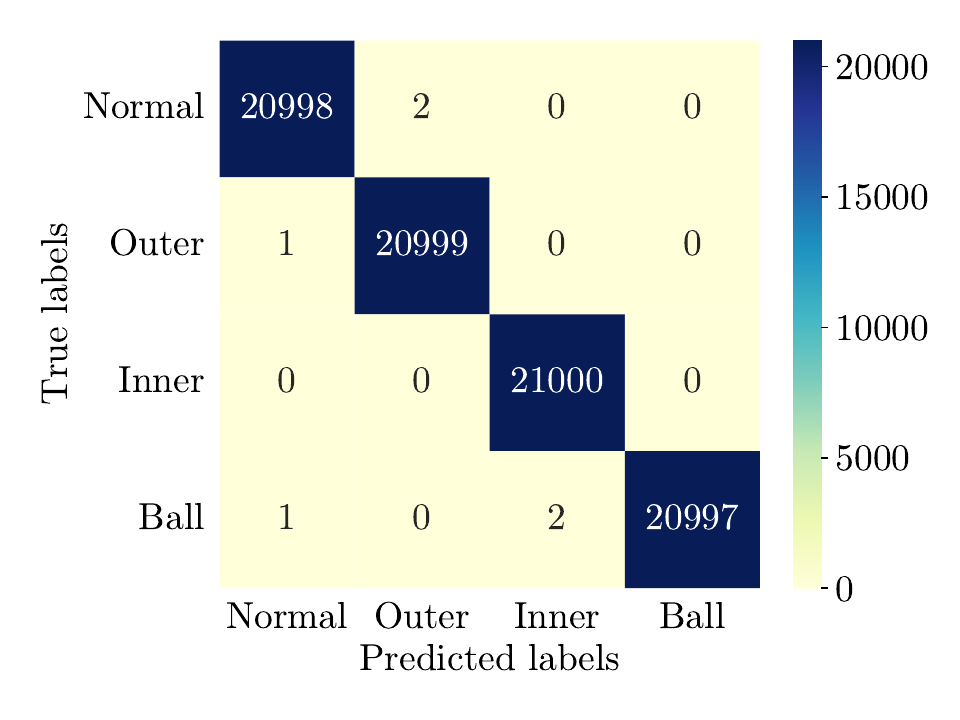}}\,
   \subfloat[SNR = 5 dB. \label{fig:cm_2}]{\includegraphics[width=.24\textwidth]{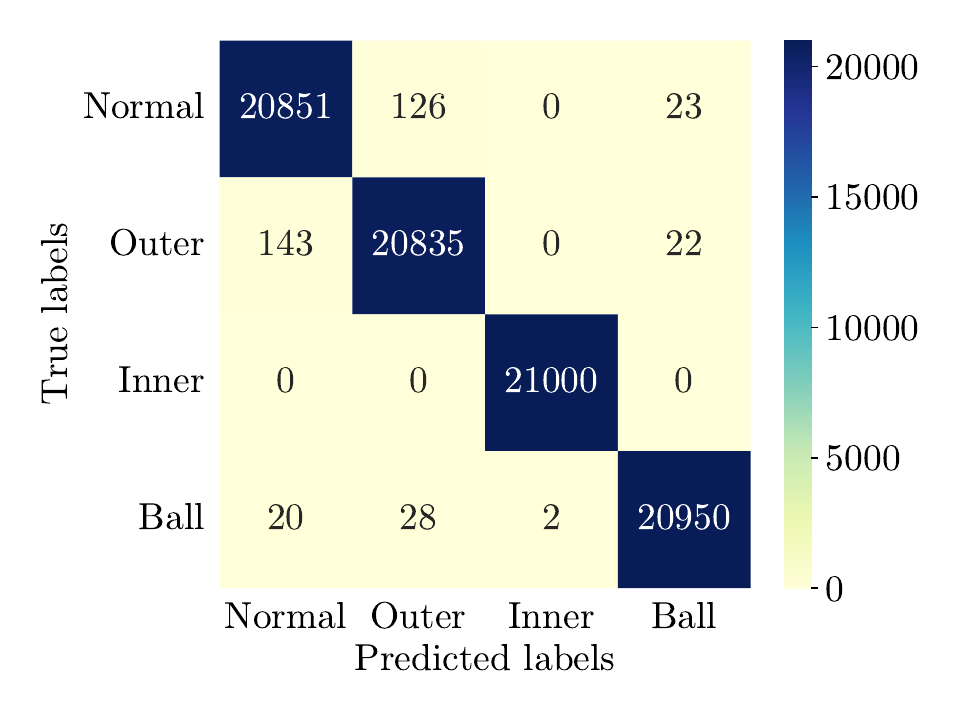}}\,
   \subfloat[SNR = 0 dB. \label{fig:cm_3}]{\includegraphics[width=.24\textwidth]{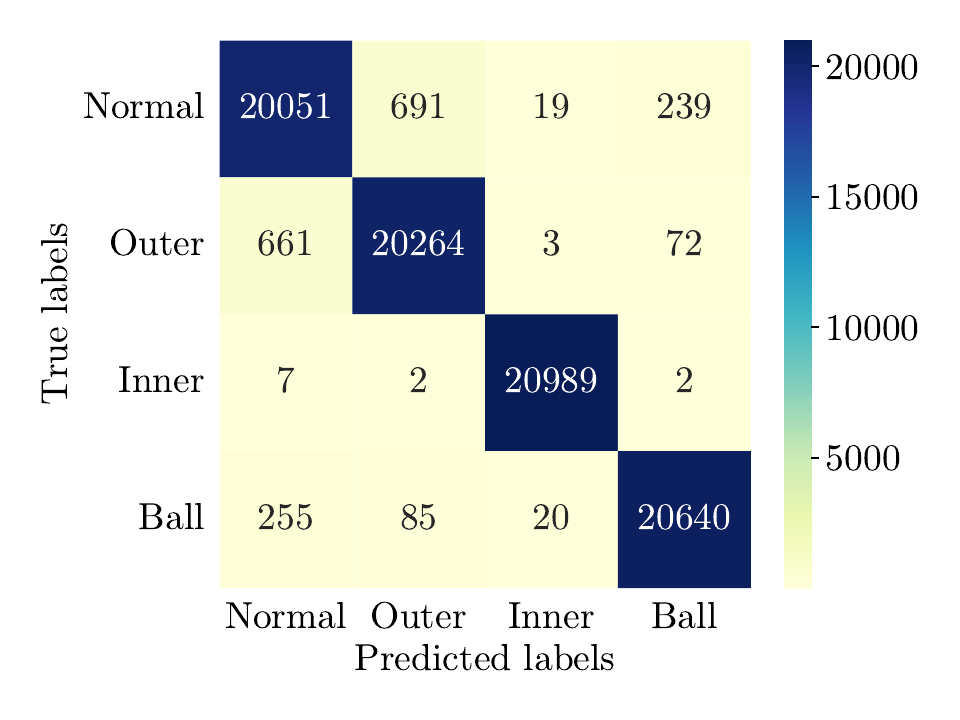}}\,
   \subfloat[SNR = -5 dB. \label{fig:cm_4}]{\includegraphics[width=.24\textwidth]{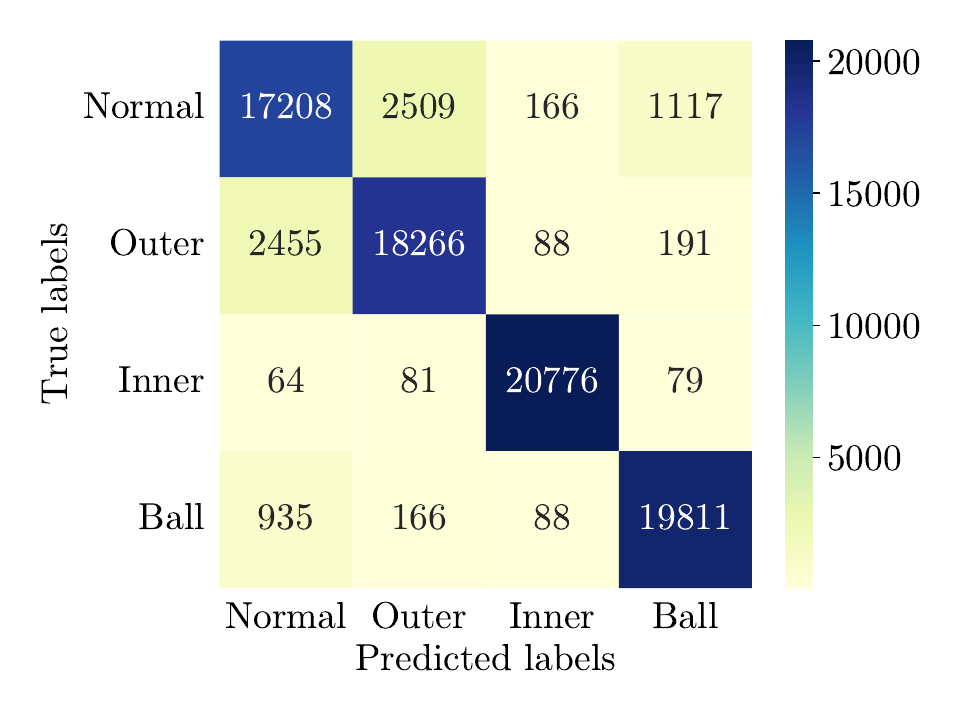}}
   \caption{The testing confusion matrices from all folds when dealing with (a) clean, and (b)-(d) noisy vibration signals.}
   \label{fig:cms}
\end{figure*}

Table \ref{table:tab_perf_results} illustrates the testing performance of the TF-CNN model and compares it to the two competing methods (T-CNN \cite{jalonen2023bearing_fault} and PIResNet \cite{Ni2023}) across different SNR levels. The results reveal a consistent superiority of the proposed model over the other techniques across all metrics and SNR levels. Additionally, they show that the boost in performance becomes more pronounced when the level of noise increases in the vibration signal. For instance, when utilizing clean data, the TF-CNN outruns PIResNet and T-CNN with 2.2\% and 1.6\% point gains in accuracy, respectively. Nonetheless, these gains increase to approximately 19\% and 15\% when dealing with severely degraded vibration signals at -5 dB. This anticipated improvement in performance can be attributed to the uniform distribution of noise within the TF lattice. Consequently, quadratic TFRs are well-suited to unveil distinctive patterns associated with the different bearing faults while concurrently mitigating the influence of noise.
\\
Moreover, the TF-CNN's confusion matrices are presented in Fig. \ref{fig:cms} and suggest that:
(1) the \emph{Inner} fault state is the simplest to predict as it has the least drop in successful predictions; from 21,000 for clean signals to 20,776 for noisy signals at -5 dB; 
(2) the \emph{Normal} and \emph{Outer} classes interlace at -5 dB because heavy noise masks their distinct TF signatures; 
(3) a mild misclassification exists at -5 dB among the \emph{Normal} and \emph{Ball} classes; 
and (4) by analyzing the confusion matrices' progression through the increasing levels of noise, there is a slight performance imbalance among the classes at -5 dB. Specifically, the \emph{Normal} and \emph{Outer} states have less successful predictions when compared to the other classes. This observation suggests that intricate TF details are required to distinguish these two states.
\begin{table}[!t]
\caption{The macro-averaged testing performance. The results are 5-fold averaged ($\pm$ standard deviations), and the best outcomes are in bold. The PIResNet results are drawn from the reported single-trial best-performing confusion matrices in \cite{Ni2023}; hence, some standard deviations are missing.}
\label{table:tab_perf_results}
\resizebox{\linewidth}{!}{
\setlength\tabcolsep{3pt} 
\begin{tabular}{cccccc}
\textbf{Model} & \textbf{SNR} & \textbf{Accuracy} & \textbf{Precision} & \textbf{Recall} & \textbf{F1-Score}
\\\toprule
PIResNet \cite{Ni2023} & Clean & 97.78 ± 0.18 & 98.0 ± N/A   & 98.0 ± N/A   & 98.0 ± N/A   \\
T-CNN \cite{jalonen2023bearing_fault}      & Clean & 98.4 ± 0.3   & 98.4 ± 0.3   & 98.4 ± 0.3   & 98.4 ± 0.3   \\
TF-CNN   & Clean & \textbf{99.99 ± 0.01} & \textbf{99.99 ± 0.01} & \textbf{99.99 ± 0.01} & \textbf{99.99 ± 0.01}
\\\midrule
PIResNet \cite{Ni2023} & 5 dB  & 90.26 ± 0.18 & 90.5 ± N/A   & 90.5 ± N/A   & 90.5 ± N/A   \\
T-CNN \cite{jalonen2023bearing_fault}      & 5 dB  & 91.2 ± 0.3   & 91.3 ± 0.4   & 91.2 ± 0.4   & 91.2 ± 0.4   \\
TF-CNN   & 5 dB  & \textbf{99.57 ± 0.05} & \textbf{99.57 ± 0.05} & \textbf{99.57 ± 0.05} & \textbf{99.57 ± 0.05}
\\\midrule
PIResNet \cite{Ni2023} & 0 dB  & 81.28 ± 0.36 & 81.6 ± N/A   & 81.6 ± N/A   & 81.6 ± N/A   \\
T-CNN \cite{jalonen2023bearing_fault}      & 0 dB  & 84.0 ± 0.3   & 84.0 ± 0.2   & 84.0 ± 0.3   & 84.0 ± 0.3   \\
TF-CNN   & 0 dB  & \textbf{97.55 ± 0.12} & \textbf{97.55 ± 0.12} & \textbf{97.55 ± 0.12} & \textbf{97.55 ± 0.12}
\\\midrule
PIResNet \cite{Ni2023} & -5 dB & 71.67 ± 0.20 & 72.3 ± N/A   & 72.0 ± N/A   & 72.2 ± N/A   \\
T-CNN \cite{jalonen2023bearing_fault}     & -5 dB & 75.3 ± 0.5   & 75.1 ± 0.7   & 75.3 ± 0.6   & 75.2 ± 0.7   \\
TF-CNN   & -5 dB & \textbf{90.55 ± 0.16} & \textbf{90.51 ± 0.17} & \textbf{90.55 ± 0.17} & \textbf{90.52 ± 0.17}
\\\bottomrule
\end{tabular}}
\end{table}
\subsection{The t-SNE and Grad-CAM analysis}
\begin{figure*}[!t]
   \centering
   \subfloat[Clean data (no added noise). \label{fig:t-sne_1}]{\includegraphics[width=.24\textwidth]{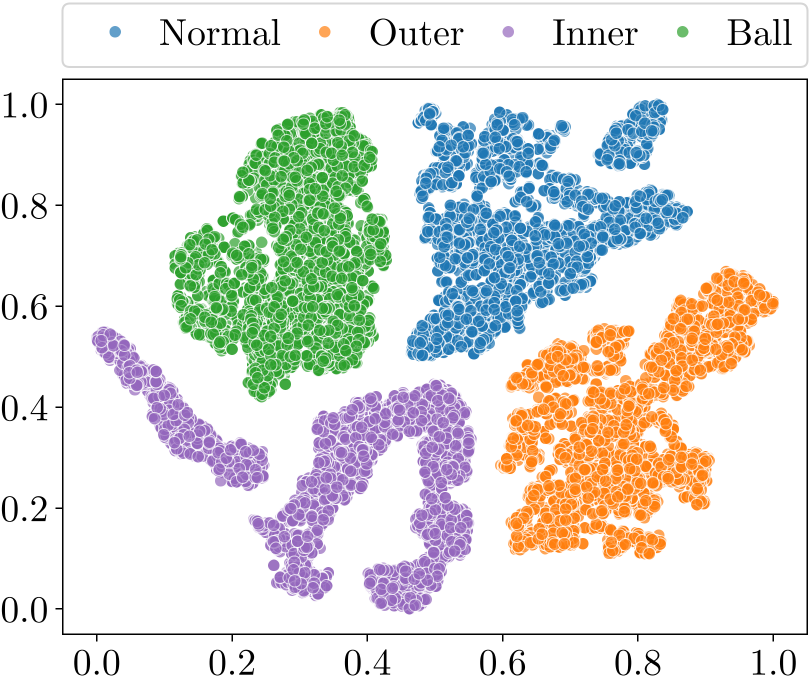}}\,
   \subfloat[SNR = 5 dB. \label{fig:t-sne_2}]{\includegraphics[width=.24\textwidth]{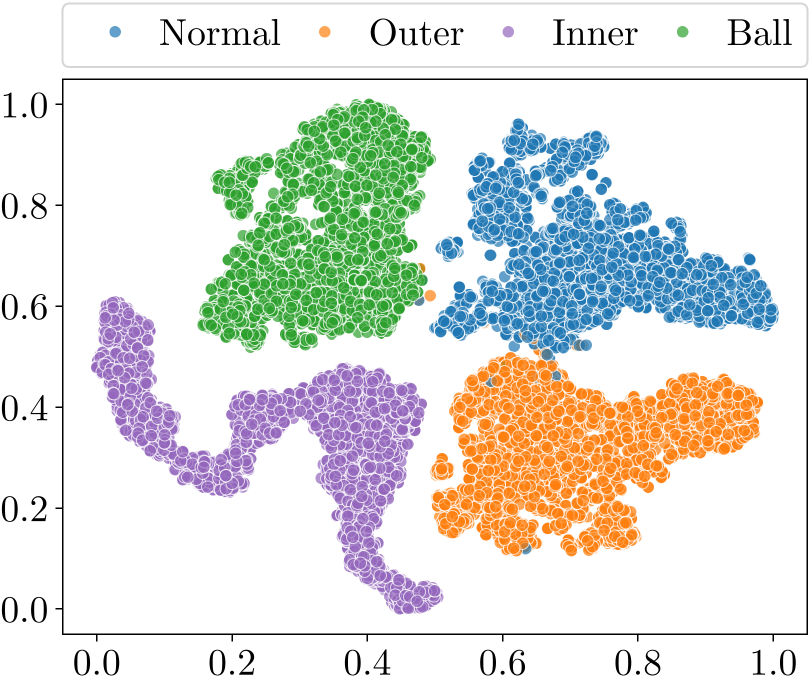}}\,
   \subfloat[SNR = 0 dB. \label{fig:t-sne_3}]{\includegraphics[width=.24\textwidth]{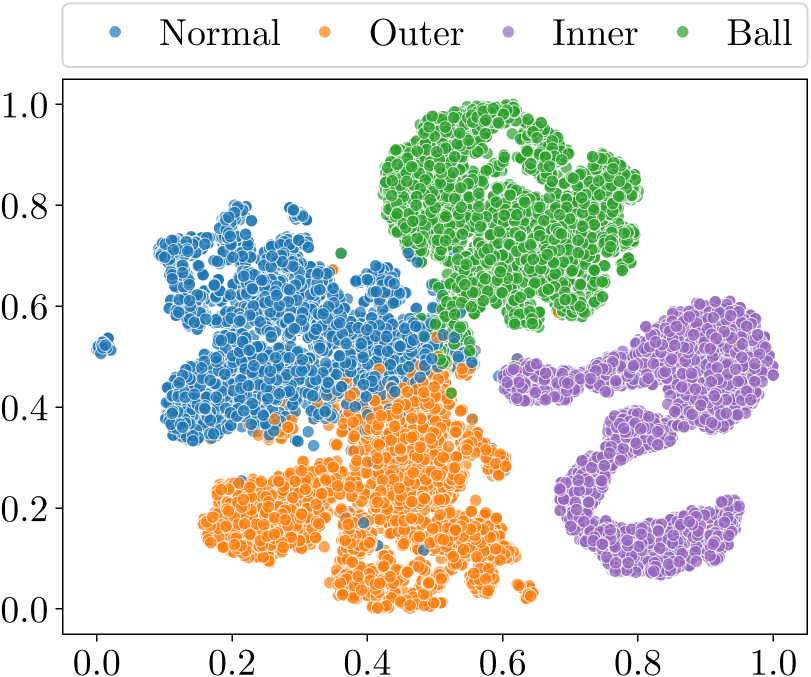}}\,
   \subfloat[SNR = -5 dB. \label{fig:t-sne_4}]{\includegraphics[width=.24\textwidth]{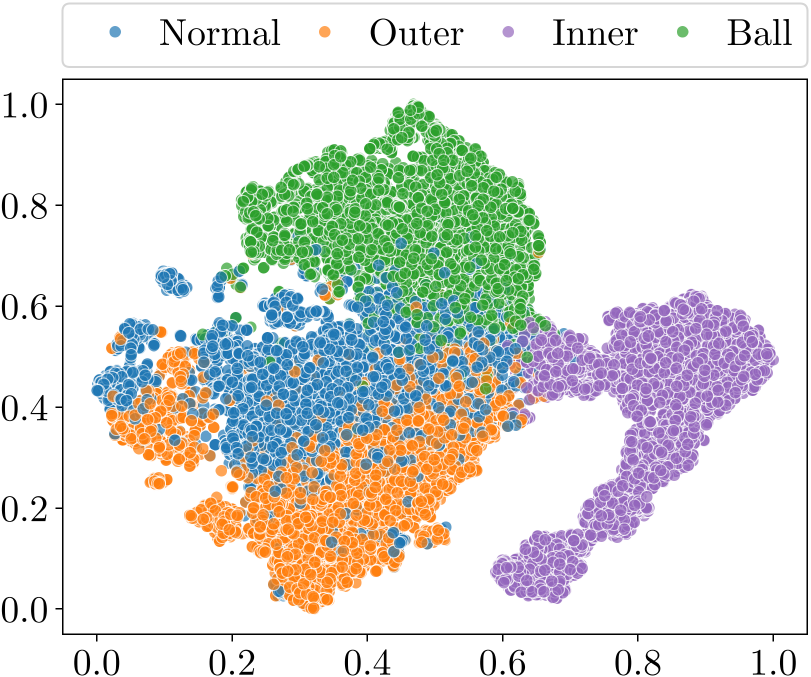}}
   \caption{The t-SNE analysis of the first testing set, using the model's last layer features, when processing (a) clean, and (b)-(d) noisy signals.}
   \label{fig:t-snes}
\end{figure*}

\begin{figure*}[!t]
    \centering
    \subfloat[Normal vibrations (\emph{Normal}). \label{fig:g1}]{\includegraphics[width=.24\textwidth]{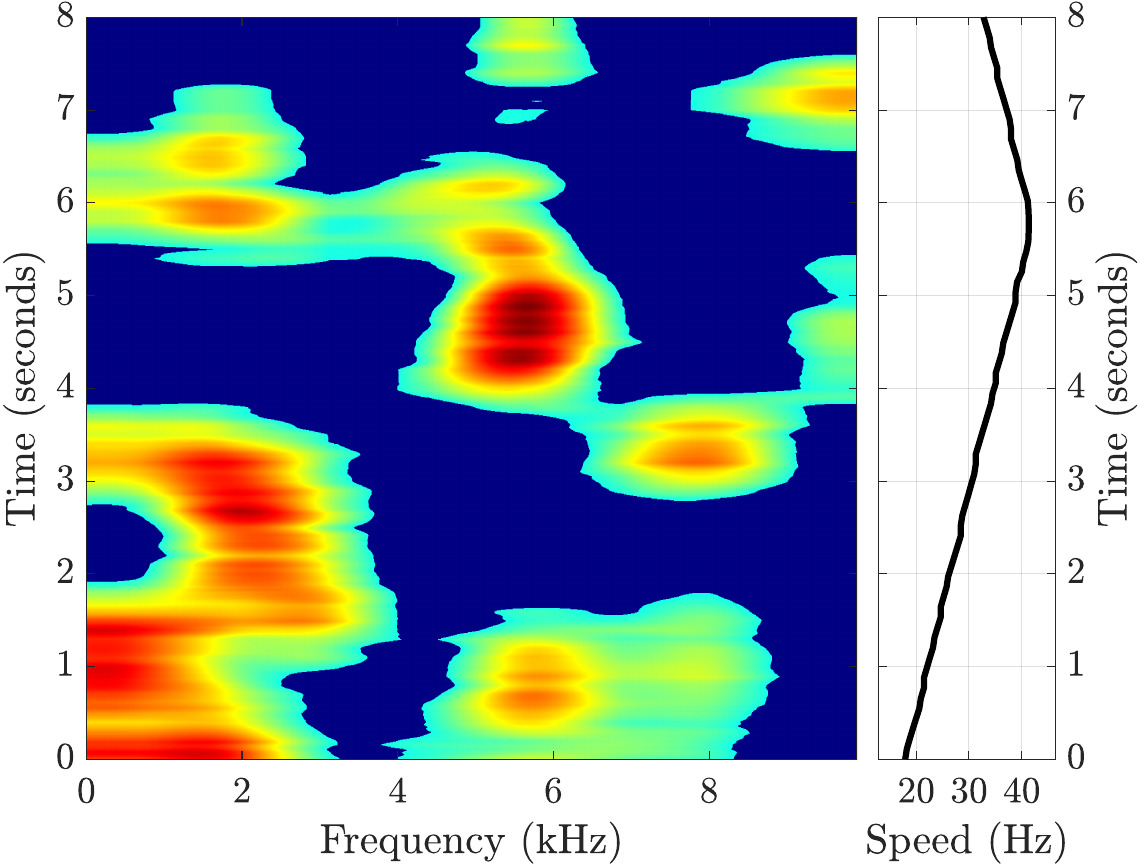}}
    \,
    \subfloat[Inner race fault (\emph{Inner}). \label{fig:g2}]{\includegraphics[width=.24\textwidth]{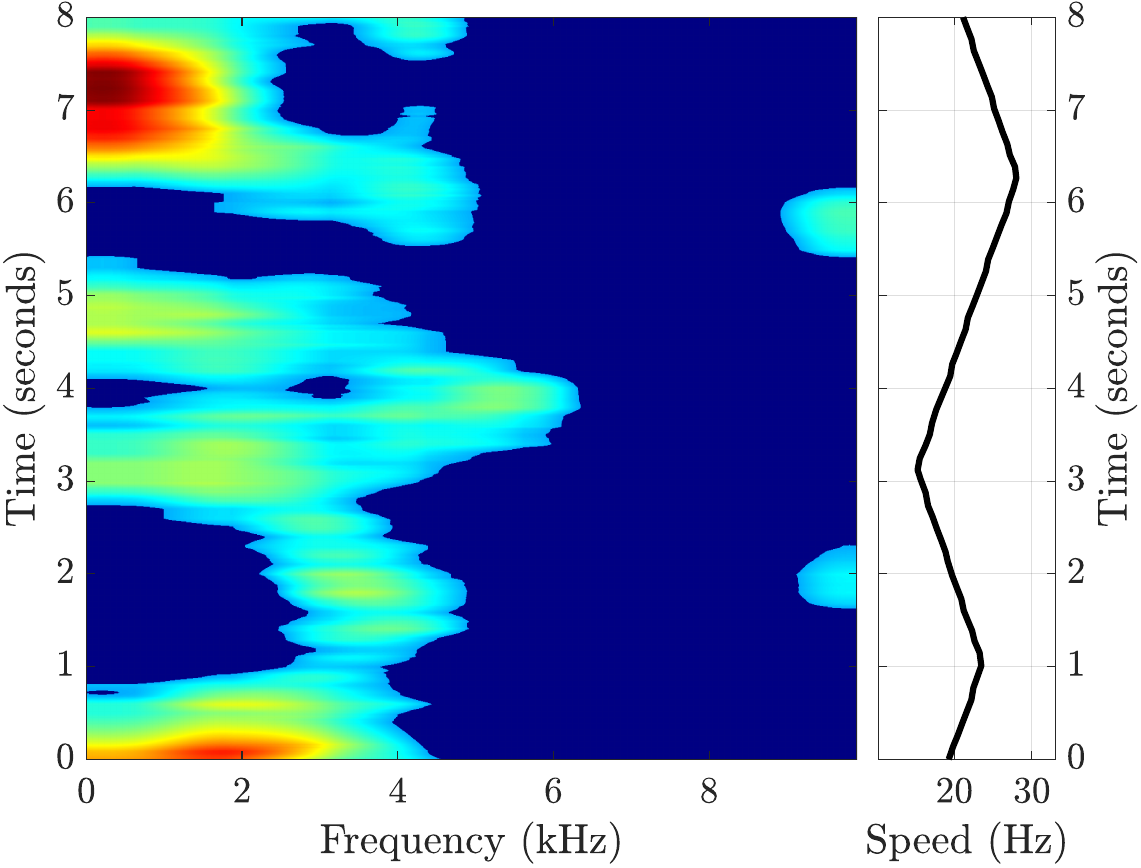}}
    \,
    \subfloat[Outer race fault (\emph{Outer}). \label{fig:g3}]{\includegraphics[width=.24\textwidth]{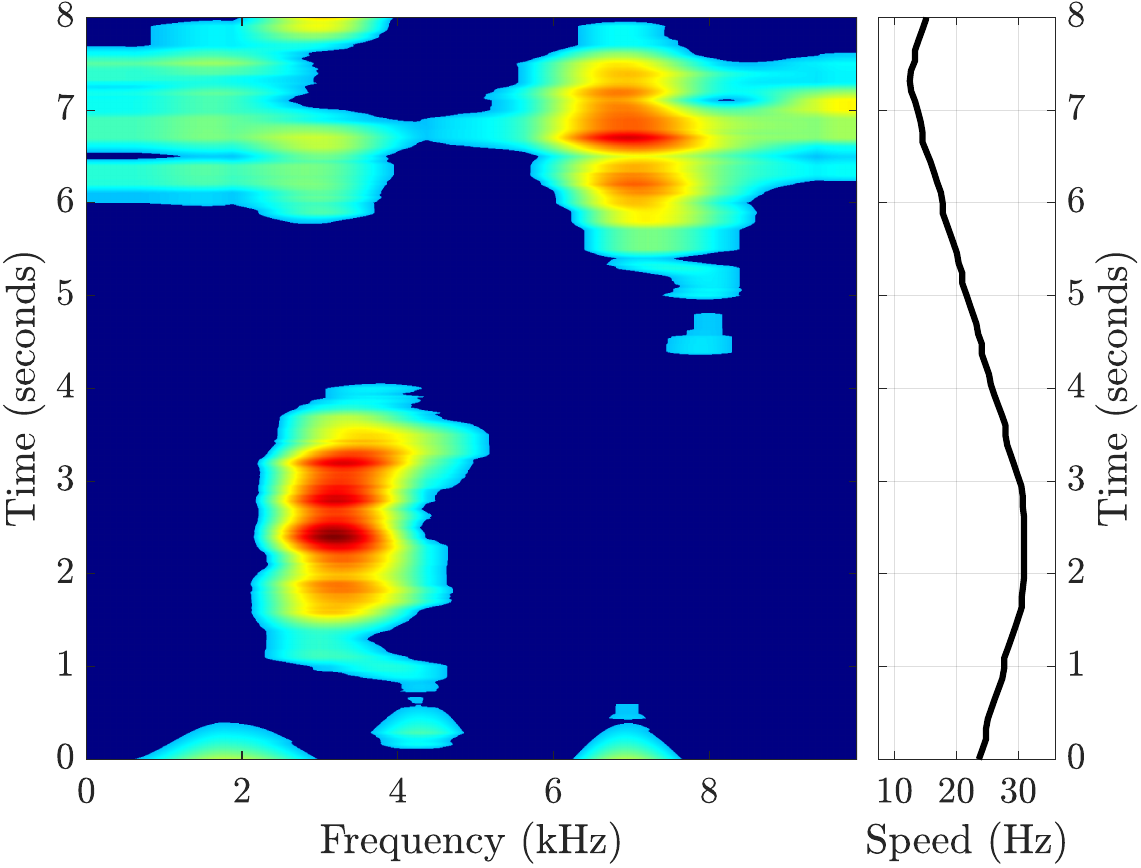}}
    \,
    \subfloat[Bearing ball fault (\emph{Ball}). \label{fig:g4}]{\includegraphics[width=.24\textwidth]{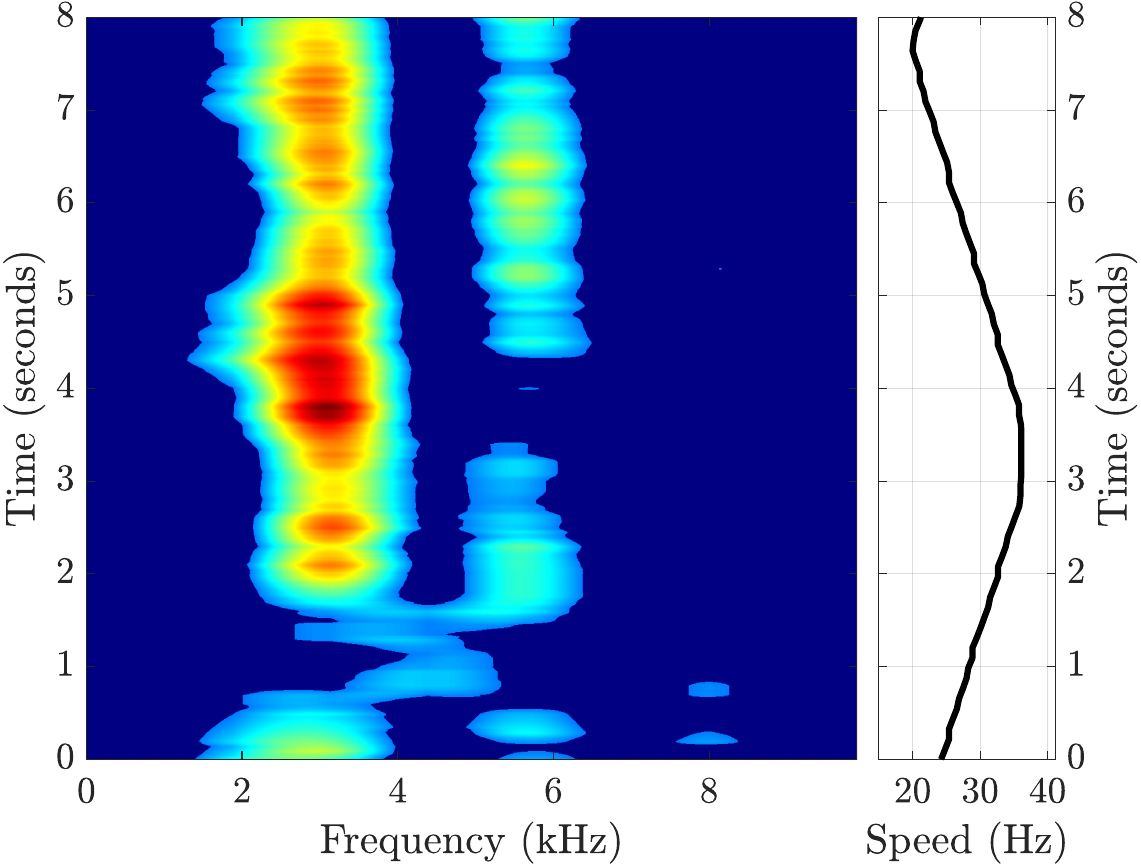}}
    \\
    \caption{The Grad-CAM maps and rotational speeds for the first 8 seconds of clean data. The results are formed as follows: (1) generate the Grad-CAM distributions for every segment (0.1 seconds of data); (2) concatenate the results of the first 80 segments in the temporal direction (8 seconds of data); (3) smooth the concatenated results with a 1-second moving average filter; and (4) threshold the smoothed distributions to show their 68.3\% confidence interval.}
   \label{fig:gradcam}
\end{figure*}

A two-dimensional representation of the TF-CNN learned features is illustrated in Fig. \ref{fig:t-snes} using t-SNE.
The results validate the model's ability to learn state-relevant features in case of clean or noisy vibration data. Specifically, Figs. \ref{fig:t-sne_1} and \ref{fig:t-sne_2} show clear inter-class separation among all classes while Fig. \ref{fig:t-sne_3} depicts a slight overlap between the \emph{Normal} and \emph{Outer} classes. Nonetheless, this overlap escalates when heavy noise is introduced as shown in Fig. \ref{fig:t-sne_4}. Moreover, the t-SNE results reveal that the \emph{Normal} and \emph{Outer} states are the most alike when degraded with heavy noise, while the \emph{Inner} and \emph{Ball} classes are the most distinct from the rest. These observations are aligned with the findings in Fig. \ref{fig:cms} and justify the slight performance imbalance at -5 dB. We hypothesize that this alikeness is linked to the characteristic frequency of outer race faults being the lowest among the faults observed. This particular frequency characteristic causes their harmonics to sway less from the resonance of the bearing. Consequently, in the TF domain, outer race faults tend to overlap more with the normal state, and substantial noise can significantly obscure their distinguishing marks.
\\
Furthermore, Fig. \ref{fig:gradcam} illustrates the Grad-CAM outcomes for all the classes along with the rotational speeds for the first 8 seconds of clean vibrations. The visualizations validate our claim that the diagnosis of bearing faults is beyond the scope of basic harmonic analysis. In particular, they underscore the necessity for dynamic features derived from the joint TF domain and indicate apparent limitations when using time-only or frequency-only representations. Additionally, the results showcase the versatility of the proposed model in acquiring fault-relevant non-stationary features, i.e., important descriptors are extracted from regions that evolve in time and frequency; see Figs. \ref{fig:g2}-\ref{fig:g4}. Finally, by associating the rotational speed profiles with the highlighted TF regions, one notes a moderate correspondence between the region's frequency support and the motor's speed, i.e., the regions are temporally coupled with the rotational speeds. However, the outcomes also demonstrate a need for improving the proposed model's attention to generate predictions with higher correlations to the operating variable speeds.
\subsection{The misclassifications link to speed}
\begin{figure}[!t]
    \centering
    \subfloat[The misclassification temporal locale w.r.t. speed. \label{fig:h1}]{\includegraphics[width=.475\textwidth]{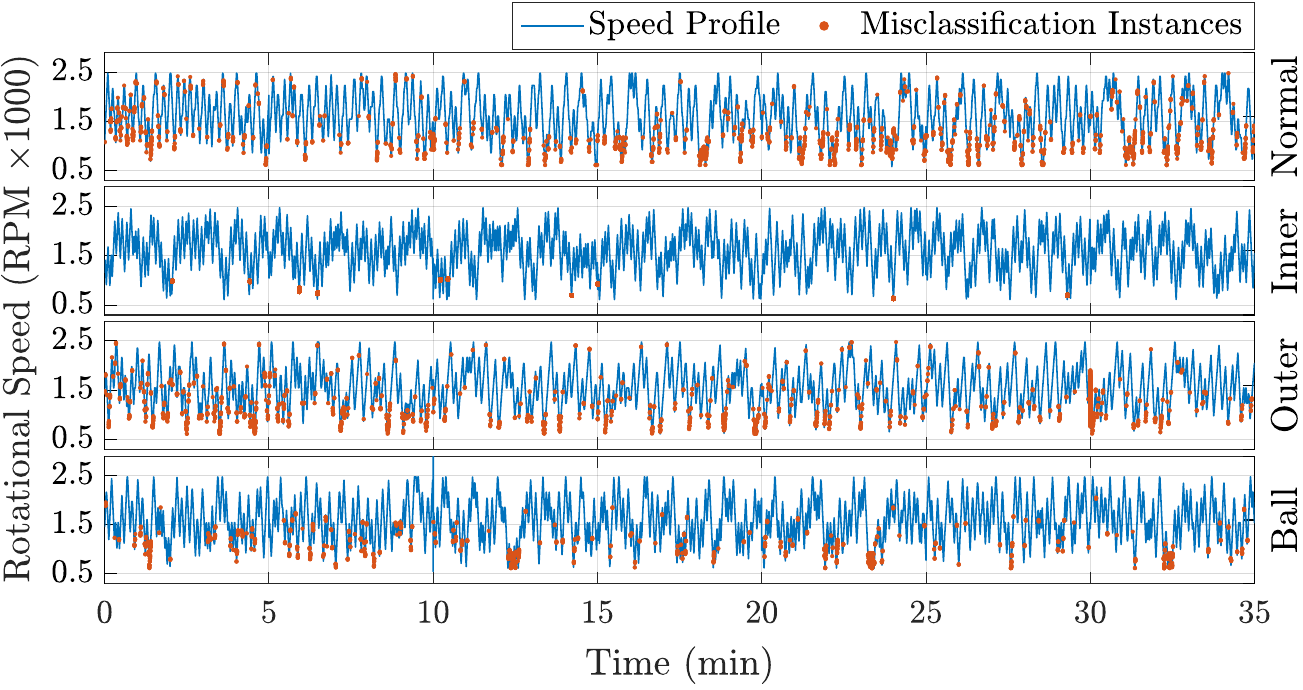}}
    \\
    \subfloat[The probability distribution of rotational speeds given misclassification. \label{fig:h2}]{\includegraphics[width=.475\textwidth]{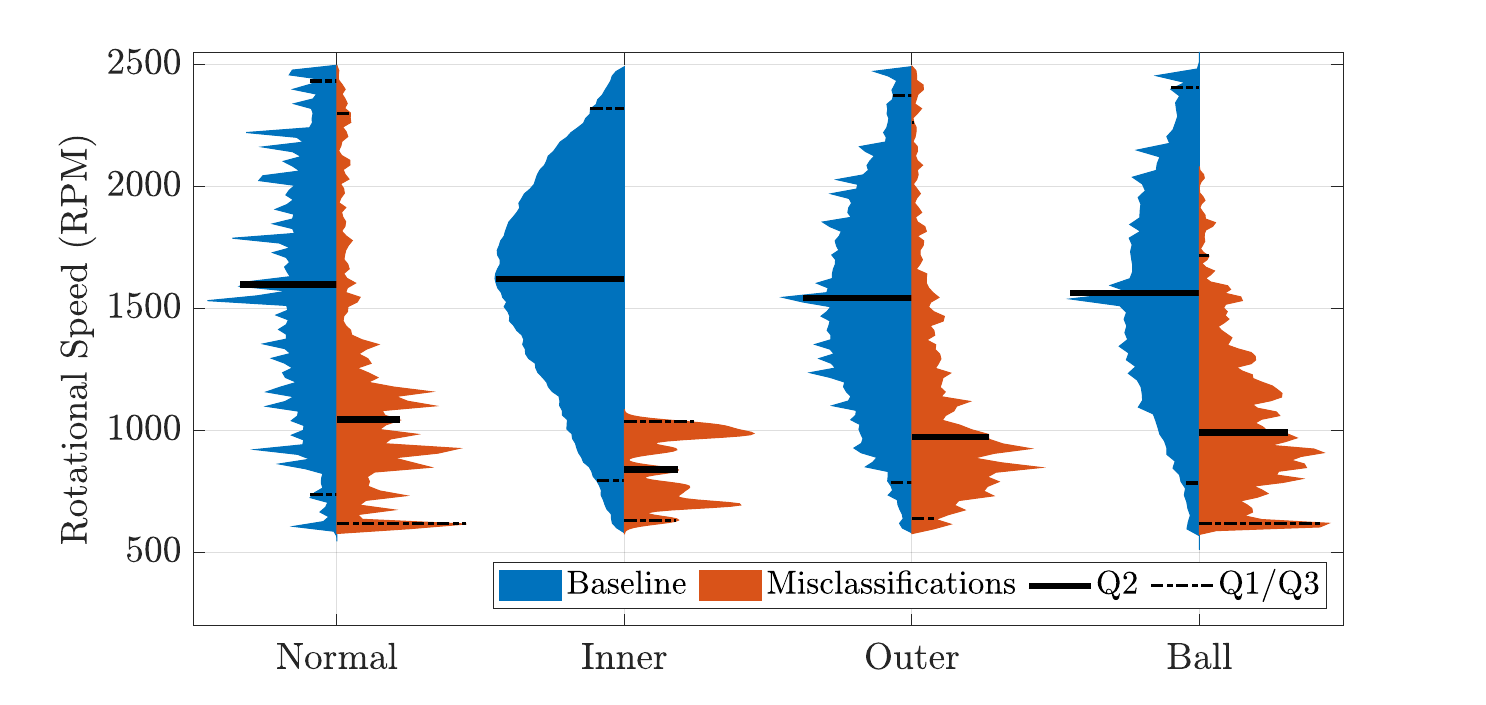}}
    \caption{The link between misclassification, at 0 dB SNR, and rotational speed for the different bearing health states. (a) shows the temporal locale of misclassifications in relation to the motor's speed profile, and (b) compares the speed conditional probability distributions given false decisions to the baseline, i.e., the unconditional probability distribution of the time-varying speed. The baseline speed profiles in (a) are interpolated by the modified Akima cubic Hermite interpolation method to match the decisions' sampling time. The probability distributions in (b) are smoothed using normal kernels and plotted along with their 25\% (Q1), 50\% (Q2), and 75\% (Q3) percentiles.}
   \label{fig:stat_results}
\end{figure}

The connection between the TF-CNN false decisions at 0 dB SNR and the motor rotational speed is explored in Fig. \ref{fig:stat_results}. First, the temporal locality of the model's misclassifications is presented in Fig. \ref{fig:h1} for all the classes. The locale is overlaid on the rotational speed profile for the entire 35 minutes of data, i.e., all the testing data from all the folds. Moreover, Fig. \ref{fig:h2} shows the motor speed probability distribution given false predictions and compares it to the speed's unconditional distribution (baseline) for all the bearing health states. Interestingly, the results reveal that misclassifications are most likely to occur at speeds below 1,200 RPM (the lower 1/3 of the entire speed range) where 70\%, 100\%, 69\%, and 73\% of all misclassifications exist for the \emph{Normal}, \emph{Inner}, \emph{Outer}, and \emph{Ball} classes, respectively. Additionally, the conditional distributions show that the median speed for which the TF-CNN fails to predict the bearing health state correctly is 1,046 for \emph{Normal}, 840 for \emph{Inner}, 974 for \emph{Outer}, and 992 for the \emph{Ball} state.
\\
The observed skewness towards low speeds can be attributed to the spectral signature of fault-induced vibrations being very close to the bearing's resonance. In other words, the frequency of fault-induced harmonics varies less at those speeds; hence, their TFRs are more likely to be misidentified. Moreover, the constructive accumulation of fault-induced impulses, as described in Eq. (\ref{eq:fault_model}), is linearly dependent on the rotational speed, i.e., the envelope of the fault signal co-varies with the speed; see Fig. \ref{fig:model_vibration} for illustration. Consequently, the ubiquity of the fault signal becomes obscure at severe noise conditions and low operational speeds. These conjectures are aligned with the t-SNE analysis in Fig. \ref{fig:t-snes} and anticipate an increase in the fault-conditioned median speed at lower SNR levels.
We verified this prediction at -5 dB SNR and found that the median speed is 1,252 for \emph{Normal} (20\% increase), 900 for \emph{Inner} (7\% increase), 1,314 for \emph{Outer} (35\% increase), and 1,244 for the \emph{Ball} state (25\% increase).
\section{Conclusions}
Bearing faults are major irregularities in rotating machines and they are the main cause of vibration. Analyzing the vibration's morphology can provide indicators to determine the bearing's health state. However, current fault diagnosis techniques are designed to operate in highly constrained environments and neglect real-life conditions, e.g., time-varying speeds and the vibration's non-stationary nature.
This study presented a fusion between time-frequency analysis and deep learning techniques to diagnose bearing faults under highly time-varying rotational speeds and multiple noise levels. First, we modeled the bearing fault-induced vibrations in a general manner and discussed their dynamic spectral evolution. We justified the vibration's non-stationary behavior and elucidated its association with the bearing's inherent properties, including its operational parameters such as rotational speed. Moreover, we discussed time-frequency analysis using high-performing quadratic distributions and showed their effectiveness in resolving the vibration waveforms of faulty bearings in comparison to conventional tools, e.g., the Wigner-Ville distribution and the Spectrogram. After that, we designed a CNN network based on the time-frequency compact kernel distribution to diagnose various faults in rolling-element bearings.
The experimental results demonstrated the superiority of our method over recently developed techniques. For instance, they showed that quadratic time-frequency distributions are well-suited to unveil distinctive dynamic patterns associated with the different bearing faults while concurrently mitigating the influence of noise. Additionally, they emphasized the versatility of the proposed model in acquiring fault-relevant non-stationary features that are temporally coupled with the rotational speed. In brief, the proposed solution has achieved a remarkable robustness to noise and significant gains in accuracy reaching up to 15\% in severe noise conditions.

Our future research directions include testing for other faults, incorporating attention mechanisms, employing post-processing methods, and fusing time-frequency tools with more advanced deep learning techniques such as Operational Neural Networks \cite{kiranyaz2020operational,kiranyaz2021self}.
\bibliography{Sections/references}
\begin{IEEEbiography}
[{\includegraphics[width=1in,height=1.25in,clip,keepaspectratio]{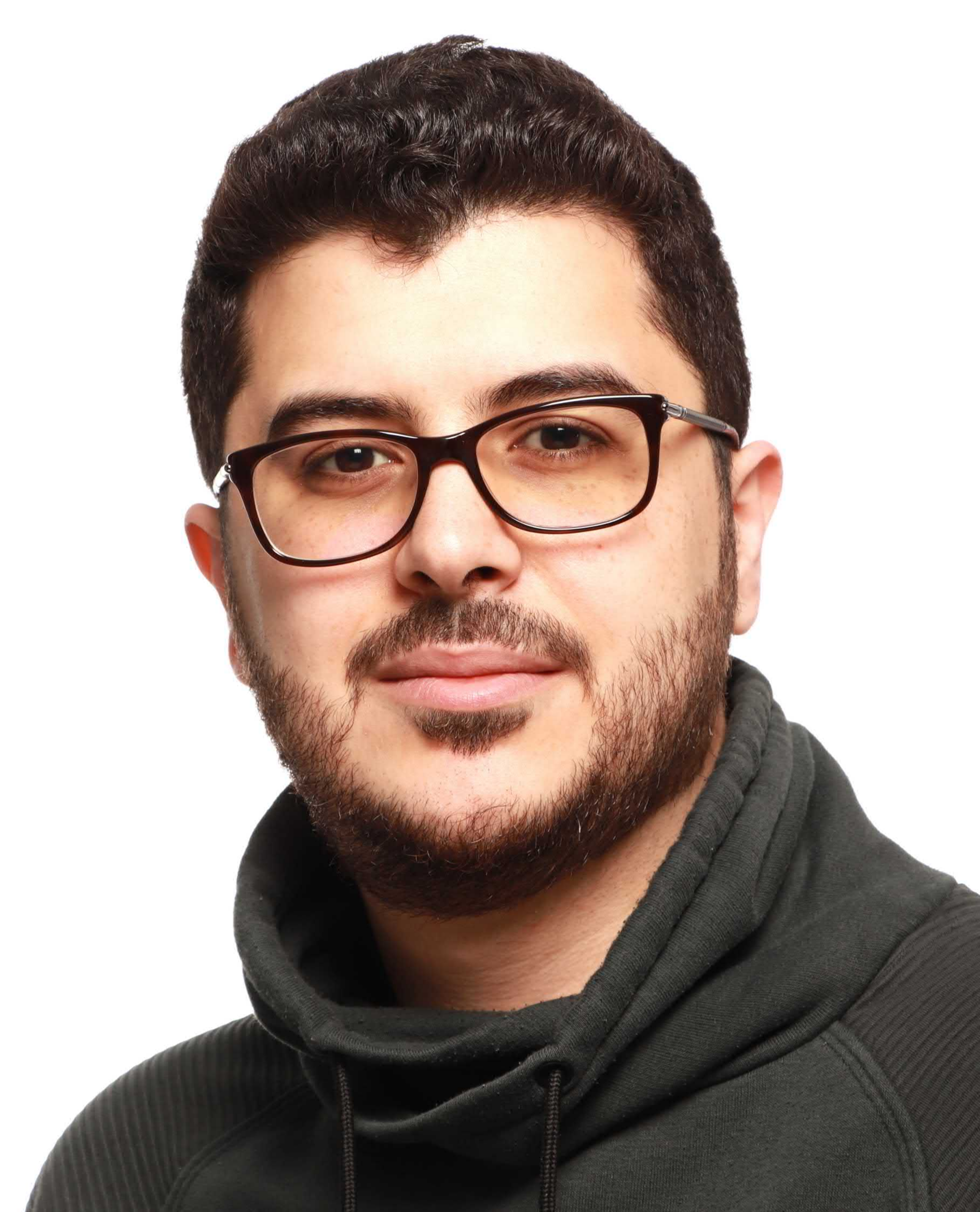}}]{Mohammad Al-Sa'd}
, Senior IEEE Member, received his B.Sc. and M.Sc. degrees in Electrical Engineering from Qatar University, Qatar, in 2012 and 2016, respectively, and his PhD degree in Electrical Engineering and Computing Sciences from Tampere University, Finland, in 2022. He specialized in signal processing and machine learning, and he is a postdoctoral fellow at the Faculty of Medicine and the Department of Clinical Neurophysiology at the University of Helsinki and Helsinki University Hospital, and at the Department of Computing Sciences at Tampere University, Finland. He has served as a technical reviewer for several journals, including IEEE Transactions on Signal Processing, IEEE Transactions on Instrumentation and Measurement, Digital Signal Processing, Signal Processing, Biomedical Signal Processing and Control, and IEEE Access. His research interests include time-frequency signal analysis, machine learning, electroencephalogram analysis and processing, information flow and theory, signal modeling, and optimization.
\end{IEEEbiography}
\begin{IEEEbiography}[{\includegraphics[width=1in,height=1.25in,clip,keepaspectratio]{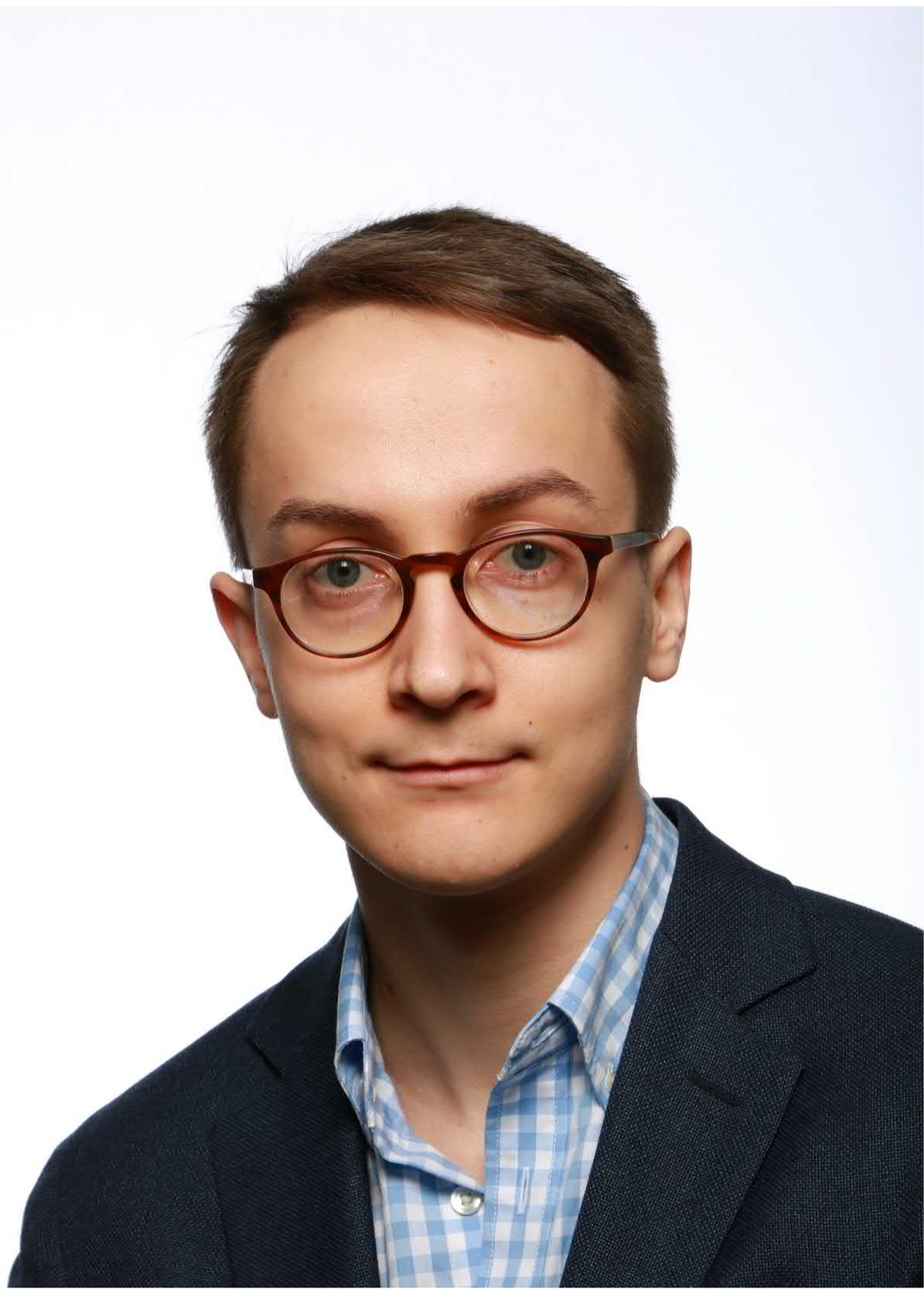}}]{Tuomas Jalonen}
received the B.Sc. and M.Sc. degrees in mechanical engineering from Tampere University, Finland, in 2017 and 2019, respectively. He is currently pursuing the Ph.D. degree with the Faculty of Information Technology and Communication Sciences of Tampere University, Finland. His research interest includes the development of machine learning applications for manufacturing.
\end{IEEEbiography}
\begin{IEEEbiography}[{\includegraphics[width=1in,height=1.25in,clip,keepaspectratio]{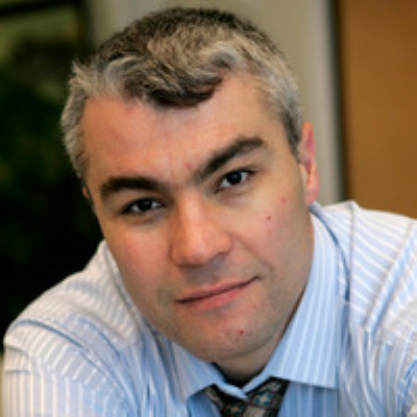}}]{Serkan Kiranyaz}
(Senior Member, IEEE) is a Professor with Qatar University, Doha, Qatar. He published two books, five book chapters, more than 80 journal articles in high impact journals, and 100 articles in international conferences. He made contributions on evolutionary optimization, machine learning, bio-signal analysis, computer vision with applications to recognition, classification, and signal processing. He has coauthored the articles which have nominated or received the “Best Paper Award” in ICIP 2013, ICPR 2014, ICIP 2015, and IEEE Transactions on Signal Processing (TSP) 2018. He had the most-popular articles in the years 2010 and 2016, and most-cited article in 2018 in IEEE Transactions on Biomedical Engineering. From 2010 to 2015, he authored the 4th most-cited article of the Neural Networks journal. His research team has won the second and first places in PhysioNet Grand Challenges 2016 and 2017, among 48 and 75 international teams, respectively. His theoretical contributions to advance the current state of the art in modeling and representation, targeting high long-term impact, while algorithmic, system level design and implementation issues target medium and long-term challenges for the next five to ten years. He in particular aims at investigating scientific questions and inventing cutting-edge solutions in “personalized biomedicine” which is in one of the most dynamic areas where science combines with technology to produce efficient signal and information processing systems.
\end{IEEEbiography}
\begin{IEEEbiography}[{\includegraphics[width=1in,height=1.25in,clip,keepaspectratio]{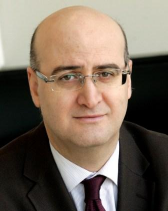}}]{Moncef Gabbouj}
(Fellow Member, IEEE) received the B.S. degree from Oklahoma State University, Stillwater, OK, USA, in 1985, and the M.S. and Ph.D. degrees from Purdue University, in 1986 and 1989, respectively, all in electrical engineering. He is a Professor of signal processing with the Department of Computing Sciences, Tampere University, Tampere, Finland. He was an Academy of Finland Professor from 2011 to 2015. His research interests include big data analytics, multimedia content-based analysis, indexing and retrieval, artificial intelligence, machine learning, pattern recognition, nonlinear signal and image processing and analysis, voice conversion, and video processing and coding. Dr. Gabbouj is a member of the Academia Europaea and the Finnish Academy of Science and Letters. He is the past Chairman of the IEEE CAS TC on DSP and the Committee Member of the IEEE Fourier Award for Signal Processing. He served as an Associate Editor and the Guest Editor of many IEEE, and international journals and a Distinguished Lecturer for the IEEE CASS. He is the Finland Site Director of the NSF IUCRC funded Center for Visual and Decision Informatics (CVDI) and leads the Artificial Intelligence Research Task Force of the Ministry of Economic Affairs and Employment funded Research Alliance on Autonomous Systems (RAAS).
\end{IEEEbiography}
\end{document}